\pdfoutput=1
\PassOptionsToPackage{most,breakable}{tcolorbox}
\PassOptionsToPackage{dvipsnames,table}{xcolor}
\PassOptionsToPackage{numbers,sort&compress}{natbib}
\documentclass[11pt]{article}

\usepackage[preprint]{ropedia}

\usepackage[T1]{fontenc}
\usepackage[utf8]{inputenc}
\usepackage{amsmath,amssymb}
\usepackage{amsfonts}
\usepackage{mathpazo}
\usepackage[xcharter,bigdelims,vvarbb]{newtxmath}
\usepackage[scaled]{helvet}
\usepackage[scaled=1.1]{zlmtt}
\usepackage{latexsym}
\usepackage{pifont}
\usepackage{placeins}
\usepackage{booktabs}
\usepackage{multirow}
\usepackage{tabularx}
\usepackage{subcaption}
\usepackage[export]{adjustbox}
\usepackage{makecell}
\usepackage{enumitem}
\usepackage{rotating}
\usepackage{titletoc}
\usepackage{comment}
\usepackage{wrapfig}
\usepackage{needspace}
\usepackage{nicefrac}
\usepackage{microtype}
\usepackage{siunitx}
\usepackage{dblfloatfix}
\usepackage{graphicx}
\usepackage{cuted}
\usepackage{capt-of}
\usepackage{xspace}
\usepackage{float}
\usepackage{listings}
\usepackage{xurl}
\setlist[itemize]{leftmargin=2em}

\usepackage{tikz}
\usetikzlibrary{fadings}
\usepackage{array}
\usepackage{tcolorbox}
\usepackage{threeparttable}

\newcommand{\projecttitle}{Embodied Turing Machines\xspace}
\newcommand{\projectsubtitle}{Stateful Code for Robot Recursive Self-Improvement\xspace}

\definecolor{etmink}{HTML}{3D3A38}
\definecolor{etmterra}{HTML}{A65F48}
\definecolor{etmcoral}{HTML}{C8553D}
\definecolor{etmgreen}{HTML}{4E9A5B}
\definecolor{etmblue}{HTML}{5E7C88}
\definecolor{etmtan}{HTML}{8C6D4F}
\definecolor{codebg}{HTML}{F6F4EE}
\definecolor{figsage}{HTML}{DCE2D3}   
\colorlet{dahaccent}{ropediadeepgreen}
\colorlet{ropedia}{ropediagreen}

\newcommand{\cmark}{\textcolor{etmgreen}{\ding{51}}}
\newcommand{\xmark}{\textcolor{etmcoral}{\ding{55}}}
\newcommand{\pmark}{\textcolor{etmtan}{$\triangle$}}

\lstdefinestyle{etm}{
  language=Python, basicstyle=\ttfamily\small\color{etmink},
  backgroundcolor=\color{codebg}, frame=leftline, framerule=1.6pt,
  rulecolor=\color{dahaccent}, framesep=6pt, xleftmargin=8pt,
  aboveskip=5pt, belowskip=3pt, columns=fullflexible, keepspaces=true,
  showstringspaces=false, breaklines=true,
  keywordstyle=\color{etmblue}\bfseries, stringstyle=\color{etmtan},
  commentstyle=\color{etmterra}\itshape,
  morekeywords={self}, emph={memory,stage,lift_dir}, emphstyle=\color{etmterra}\bfseries}
\definecolor{codewin}{HTML}{F7F7F5}
\definecolor{codehead}{HTML}{ECECE8}
\definecolor{codedot}{HTML}{C9C9C4}
\definecolor{codenum}{HTML}{9A9A9A}
\definecolor{codeink}{HTML}{262626}
\definecolor{hlperc}{HTML}{9EBACF}
\definecolor{hltask}{HTML}{D3E6C3}
\definecolor{hlmanip}{HTML}{9DC8C7}
\definecolor{filefill}{HTML}{D3E6C3}
\definecolor{fileedge}{HTML}{6EB385}
\lstdefinestyle{etmwin}{
  language=Python, basicstyle=\ttfamily\fontsize{8.4}{11.2}\selectfont\color{codeink},
  columns=fullflexible, keepspaces=true, showstringspaces=false,
  breaklines=true, breakatwhitespace=true, breakindent=5.5em, numbers=none,
  keywordstyle={}, stringstyle={}, commentstyle=\color{codenum}\itshape, emph={}, emphstyle={},
  escapeinside={(*@}{@*)}, aboveskip=0pt, belowskip=0pt, xleftmargin=0pt, upquote=true,
  frame=none, framerule=0pt, framesep=0pt, rulecolor=\color{codewin}, backgroundcolor=\color{codewin}}
\newcommand{\lnum}[1]{\makebox[0pt][r]{\fontsize{7.4}{11.2}\selectfont\color{codenum}#1\hspace{8pt}}}

\newcommand{\codehl}[2]{{\setlength{\fboxsep}{0.7pt}\colorbox{#1}{\vphantom{Ag}#2}}}
\newcommand{\hlP}[1]{\codehl{hlperc}{#1}}     
\newcommand{\hlT}[1]{\codehl{hltask}{#1}}     
\newcommand{\hlM}[1]{\codehl{hlmanip}{#1}}    
\newcommand{\hlS}[1]{{\setlength{\fboxsep}{0.7pt}\setlength{\fboxrule}{0.5pt}%
  \fcolorbox{codeink}{codewin}{\vphantom{Ag}#1}}}     
\newcommand{\codedots}{\tikz[baseline=-0.55ex]{\foreach \i in {0,1,2}{\fill[codedot] (\i*0.16,0) circle (0.048);}}}
\newcommand{\codetab}[1]{\tikz[baseline=(t.base)]{\node[fill=codewin, rounded corners=2pt, inner xsep=5pt,
  inner ysep=3pt] (t) {\tikz[baseline=-0.55ex]{\filldraw[fill=filefill, draw=fileedge, line width=0.35pt]
  (0,-0.085) -- (0,0.085) -- (0.09,0.085) -- (0.13,0.045) -- (0.13,-0.085) -- cycle;}%
  \hspace{3pt}\ttfamily\fontsize{8.4}{10}\selectfont\color{codeink}#1};}}
\newtcblisting{codewindow}[2][]{enhanced, listing only, colback=codewin, colframe=codewin, boxrule=0pt,
  arc=3pt, left=24pt, right=6pt, top=5pt, bottom=5pt, title={\codedots\hspace{10pt}\codetab{#2}},
  colbacktitle=codehead, coltitle=codeink, toptitle=3pt, bottomtitle=0pt, lefttitle=8pt, titlerule=0pt,
  before skip=6pt, after skip=3pt, listing options={style=etmwin,#1}}
\newcommand{\codecaption}[1]{\par\noindent{\small #1\par}\vspace{6pt}}
\newcounter{codelisting}
\newcommand{\codelabel}[1]{\refstepcounter{codelisting}\label{#1}}


\colorlet{findingback}{ropedialightgreen}
\colorlet{findingbacksoft}{ropediaabstractgreen}
\colorlet{findingframe}{ropediadeepgreen}
\colorlet{findingaccent}{ropediaaccent}

\newtcolorbox{findingquote}{enhanced, left=6pt, right=6pt, top=4pt, bottom=4pt, colback=findingback,
  colframe=findingframe, coltext=ropediaink, boxrule=0pt, borderline west={1.3pt}{0pt}{findingaccent},
  arc=1pt, outer arc=1pt, before skip=6pt, after skip=6pt}
\newcommand{\finding}[2]{\begin{findingquote}\textbf{\color{ropediadeepgreen}Finding #1.}\enspace\itshape #2\end{findingquote}}
\newcommand{\advpoint}[2]{\par\medskip\noindent{\sffamily\bfseries\color{dahaccent}#1\enspace#2}\enspace}
\newcommand{\runin}[1]{\par\medskip\noindent{\sffamily\bfseries\color{dahaccent}#1}\enspace}
\definecolor{coapbg}{HTML}{E2EFD6}
\newcommand{\blocktitle}[2][etmink]{\rule{0pt}{2.5ex}{\scriptsize\sffamily\bfseries\color{#1}#2}}

\firstpagelogos{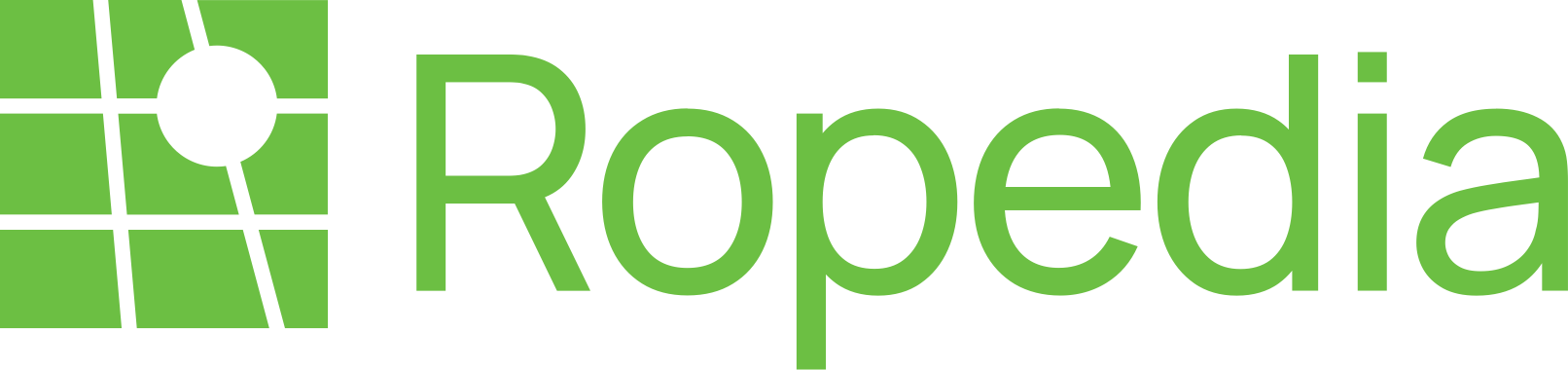}{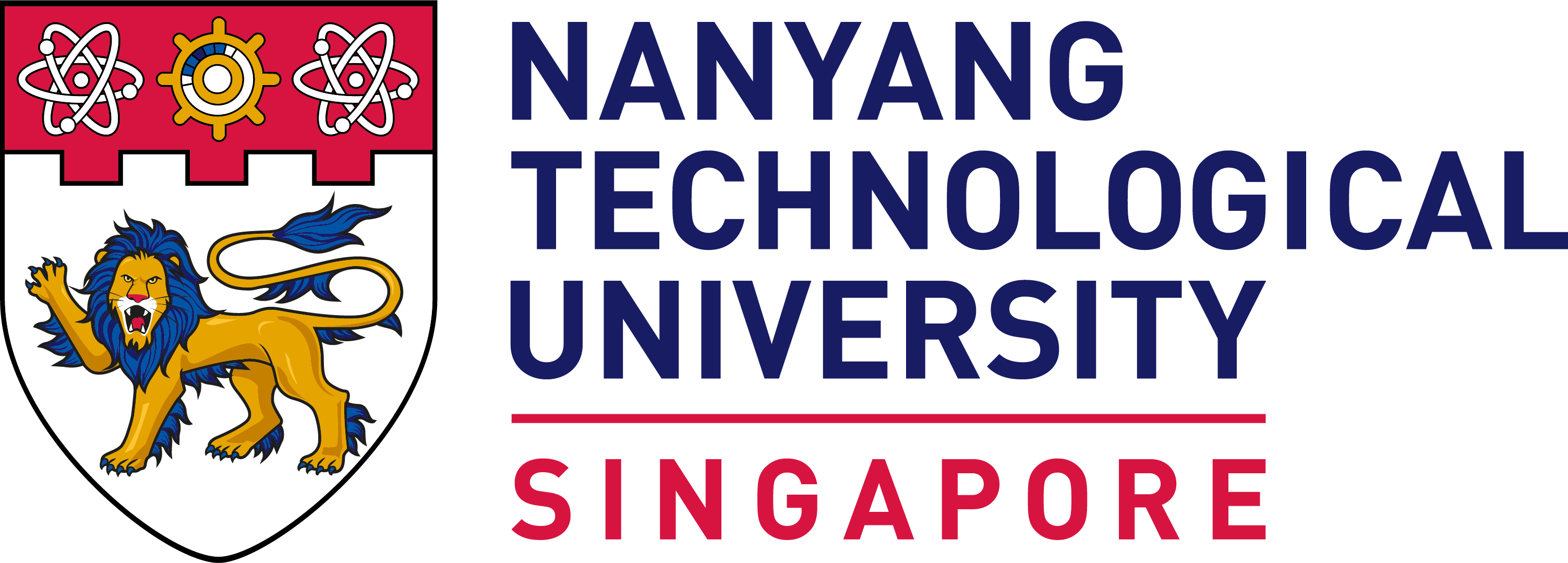}{}
\abstractlogo{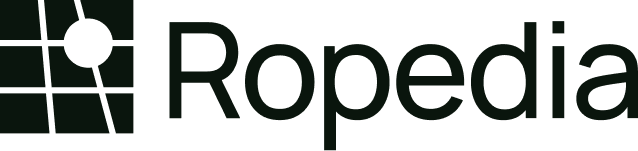}

\title{\projecttitle}
\subtitle{\projectsubtitle}

\newcommand{\authorfootnotes}{%
  \begingroup
  \renewcommand{\thefootnote}{}%
  \footnotetext{Technical Report. * Equal contribution.}%
  \endgroup
  \addtocounter{footnote}{-1}%
}

\author{
  \AuthorName{Kairui Hu}{1,2,*}\quad
  \AuthorName{Siyuan Hu}{1,*}\quad
  \AuthorName{Fangzhou Hong}{1,2}\quad
  \AuthorName{Zhaoxi Chen}{1,2}\quad
  \AuthorName{Ziwei Liu}{1,2}\\[1.0ex]
  {\normalfont\footnotesize
  \mbox{\textsuperscript{1}Nanyang Technological University}\quad
  \mbox{\textsuperscript{2}Ropedia}}
}

\begin{document}
\maketitle
\authorfootnotes

\noindent
\begin{minipage}{\textwidth}
\centering
\includegraphics[width=\linewidth]{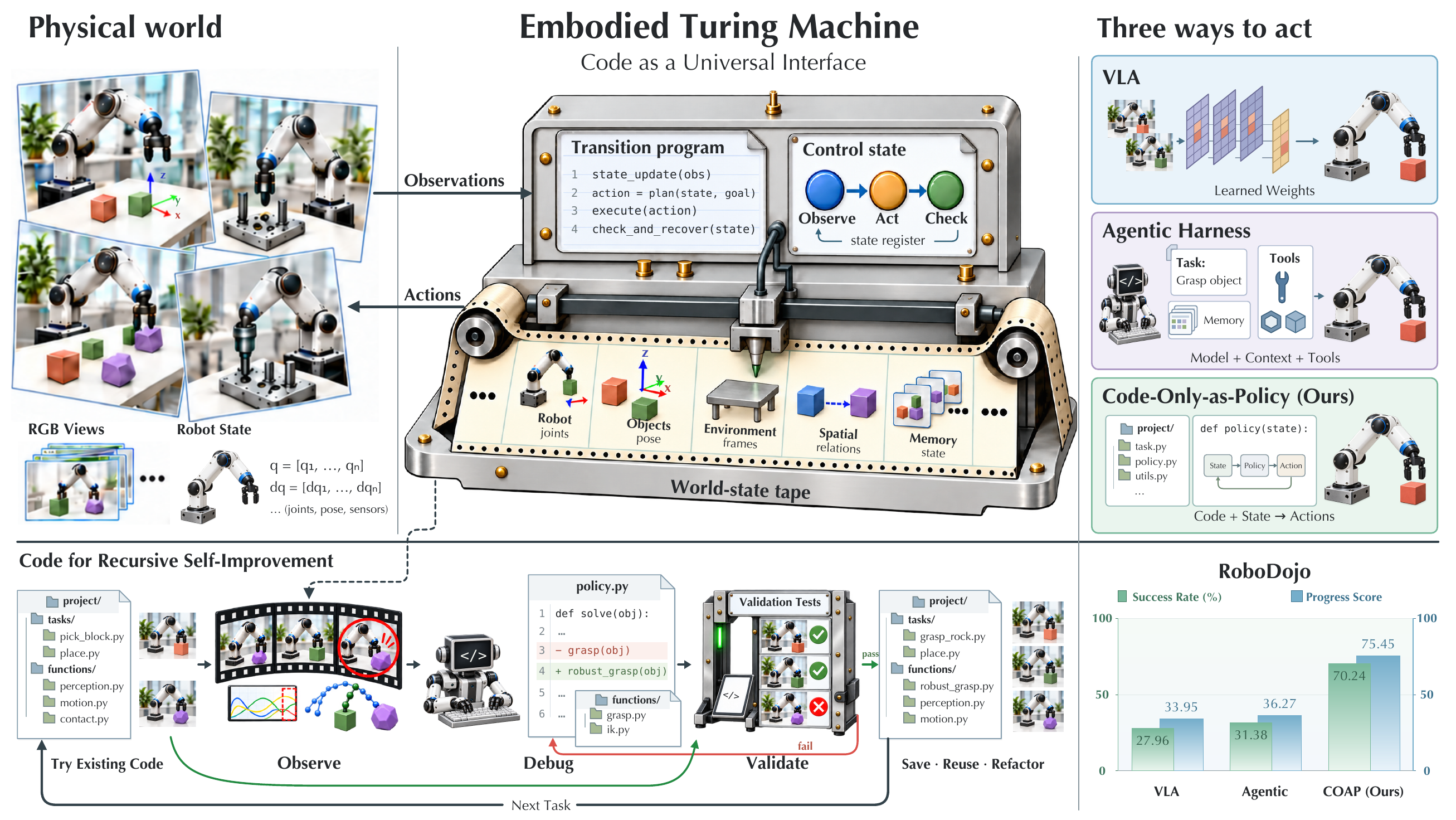}
\captionof{figure}{\textbf{Code-Only-as-Policy (COAP).} We view the embodied
world as an Embodied Turing Machine whose tape is the state of the robot and
its environment (top middle). If this state is represented accurately and the
code is robust enough, no agent is needed in the control loop (top right). COAP
is a shared library: different tasks use the same library, and within a task the
code runs across episodes. The policy is therefore code only and reaches a
70.24\% success rate on RoboDojo.}
\label{fig:teaser}
\end{minipage}
\par\vspace{8pt}

\begin{abstract}
Most robot policies keep a model in the control loop: a VLA maps
observations to actions, and an Agent Harness, such as Agent-as-Policy or Harness VLA queries a VLM for decision making
at run time. We propose a different view: the embodied world is an
\textbf{Embodied Turing Machine}, whose tape is the robot and environment state and rules are the policy. If this state can be represented
accurately, the decision making can be written entirely in code. We therefore propose
\textbf{Code-Only-as-Policy (COAP)}: code measures and tracks the robot,
environment, and task state from camera images and proprioception, and makes
every decision from it. The same code applies across episodes, and different tasks
share one library without a VLM or VLA in the loop. Compared with VLAs
and Agent Harnesses, we analyze three advantages of COAP: (i) \textbf{Explicit State}: the
state can be stored in code; (ii) \textbf{Execution}: code makes decision
making controllable, recovers from failures flexibly, and runs fast and cheaply
online; (iii) \textbf{Extensibility}: new tasks reuse, inherit, or extend the
shared library, so capabilities can accumulate over tasks. These advantages make COAP a suitable
medium for \textbf{recursive self-improvement (RSI)}: coding agents develop the
library in a closed loop, and each change is explicit and controllable. On
RoboDojo's 42 bimanual tasks, the resulting library reaches a success
rate of 70.24\% without a model at test time. The upper bound of COAP lies in how accurately the state
is represented for decision making and how robust the code logic is. We thus
propose COAP as a new paradigm for embodied tasks; since it applies across
episodes, it can also serve as an efficient data engine for VLAs and Agent
Harnesses.

\end{abstract}

\section{Introduction}
\label{sec:introduction}

\begin{flushright}
\begin{minipage}{0.66\linewidth}
\normalsize\itshape ``It is possible to invent a single machine which can be
used to compute any computable sequence.''
\par\vspace{2pt}\normalsize\upshape\hfill Alan Turing~\cite{turing1936computable}
\end{minipage}
\end{flushright}
\vspace{4pt}

A Turing machine comprises a tape that stores symbols and a set of fixed rules
that read from and write to the tape~\cite{turing1936computable}. With the
rules fixed and only the tape varying, the same rules can process arbitrary
inputs. The same abstraction applies to the physical world, where the world
state is analogous to the tape. A robot reads this tape through perception and
writes to it through action, and its policy plays the role of the rules. We
therefore model a robot and its environment as an \textbf{Embodied Turing
Machine}, whose tape holds two kinds of state: the environment state (e.g.,
object poses and their spatial relations) and the robot state (e.g., arm pose
and reachable poses).

Both kinds of state are accurately observable. The robot state is available
directly through proprioception, and object poses can be recovered from RGB
images using measured object geometry. A policy that acts
on this state gains two advantages. First, more accurate state enables more
precise action. Second, when objects are rearranged (like different episodes in a task), only the tape changes, and
the same rules (policy) still apply.

A policy can map state to action in three ways, which differ in how explicit
this mapping is. (i) \textbf{Vision-language-action models (VLAs).} A VLA
takes observations, such as camera images, as input and
learns the mapping to action, which remains implicit in its
weights~\cite{black2025pi0,kim2024openvla}.
(ii) \textbf{Agent harnesses.} The vision-language model
(VLM) either controls the robot
directly, as in Agent-as-Policy, or directs a VLA, as in
Harness VLA~\cite{chen2026showharness,zhang2026harness}. The
harness makes part of the process explicit, such as textual memory, but the
VLM and VLA remain black boxes, so we may still neither fully inspect nor control how an action is derived. 
(iii) \textbf{Code-Only-as-Policy (ours).} In contrast,
\textbf{code is a white box} that implements the Embodied Turing Machine
directly: the state, such as object poses, is stored in code variables, where
it remains accurate, persistent, and explicit; the rules are written as code
logic that computes each action from this state. With both the state and the
rules explicit, no model is required at runtime: the code is frozen and serves as
the sole policy during execution. Code thus becomes \textbf{a universal interface to the
physical world}: a single code program takes over the roles of the VLAs, the prompts and skills of Agent-as-Policy, and the memory of Harness VLA.

\begin{figure}[t]
\centering
\includegraphics[width=\linewidth]{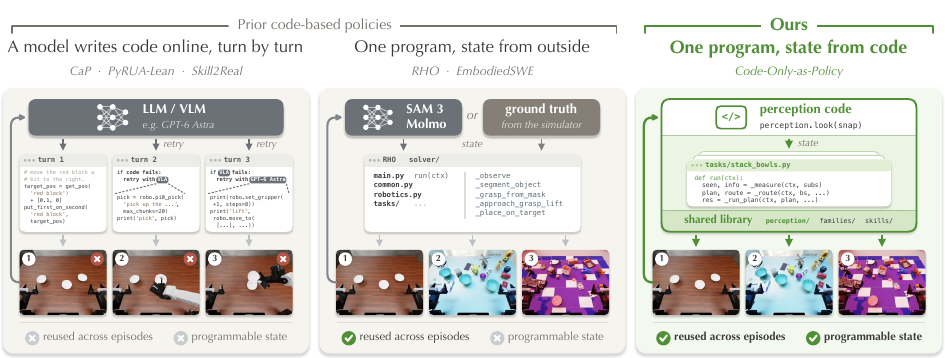}
\caption{\textbf{Comparison of code-based robot policies.} In prior work, a
model writes code-as-policy online, turn by turn, as the robot runs (left), or one
program takes its state from a learned model or the simulator oracle (middle). Our COAP reuses code across episodes and measures the state
by code (right).}
\label{fig:code_policies}
\end{figure}

Existing code-based policies either (i) ask a model to write new code-as-policy online
in each episode of a task~\cite{liang2023code,si2026pyrua,he2026skill2real},
or (ii) take the state from a model or the simulator's
oracle~\cite{elmaaroufi2026rho,shen2026embodiedswe}
(Figure~\ref{fig:code_policies}). COAP instead measures the state with its own
code and builds a shared code library for perception and execution. Different
tasks run the same library code, and only a small part of the logic is
task-specific: 83\% of the code that a new task runs comes from the library.
Nor is the code written for a particular episode: one program runs on every
episode of a task. Because the task is fixed, code that is robust enough needs
no VLM or VLA in the loop; if it still needs one, the code is not yet robust
enough to run reliably across episodes.

Code-Only-as-Policy has three advantages (\S\ref{sec:compare}).
(i) \textbf{Explicit State}: the state is stored in code variables, and every
decision is made from them (\S\ref{sec:cmp-state}). (ii) \textbf{Execution}:
the code gives fine-grained control over the state and actions, recovers from
failures flexibly by backtracking and re-attempting, and runs fast and cheaply
online (\S\ref{sec:cmp-exec}). (iii) \textbf{Extensibility}: a new task reuses,
inherits, or extends the shared library, so the code evolves quickly and
capabilities accumulate (\S\ref{sec:cmp-ext}). Hence, COAP is a suitable medium for stable recursive
self-improvement (RSI) (\S\ref{sec:dev}). Since the code is not specific to any
episode, one policy runs on different episodes and can generate successful
trajectories efficiently for VLAs and Agent Harnesses
(\S\ref{sec:dataengine}).

We evaluate COAP on RoboDojo's 42 bimanual tasks~\cite{chen2026robodojo}. A coding agent (Opus 5.5 Max) develop the code
library in a closed RSI loop that we design: under fixed rules, such as no hack
for single episodes and no oracle state at test time, the coding agent summarizes, writes,
and debugs the code without human in the loop. The development does not involve test set of a benchmark. The resulting COAP policy reaches 70.24\% success rate, 38.9\% above the state-of-the-art method. The upper bound of COAP can be pushed further if the state representation gets more accurate and code logic grows more robust. 

The weakness of COAP mainly lies in its generalization capabilities since it
needs a small amount of task-specific code logic. However, Agent-as-Policy and Harness VLA also involve
task-level memories, and some even rely on episode-level logic.
In this work, our aim is to explore the upper bound of Code-Only-as-Policy for the community.
Moreover, COAP also generates successful trajectories stably and efficiently, which can
benefit the development of VLAs and Agent Harnesses (\S\ref{sec:dataengine}). Our contributions are:
\begin{itemize}[leftmargin=1.6em]
\setlength{\itemsep}{1pt}\setlength{\parskip}{0pt}\setlength{\topsep}{2pt}
\item[(i)] \textbf{Code-Only-as-Policy.} We view the embodied world as a Turing
machine in which the robot and environment state can be
represented explicitly.
The same code applies to different episodes, and different tasks use a shared library developed offline by coding agent.
\item[(ii)] \textbf{Advantages of Code-Only-as-Policy.} 
We analyze three
advantages of code over VLAs and Agent Harnesses: \emph{Explicit State},
\emph{Execution}, and \emph{Extensibility}. We explore code-only as a new paradigm for embodied task: when the code is robust enough and applies across episodes, no VLM or VLA
is required during runtime.
\item[(iii)] \textbf{Code as the medium of RSI.} Code iterates efficiently,
and each change is controllable. Code-Only-as-Policy can also serve as an efficient data engine to
produce beneficial trajectories as new training data for VLAs, Agent-as-Policy, etc.
\end{itemize}
\FloatBarrier

\section{Related Work}
\label{sec:related}

Existing robot policies built with foundation models differ in where they keep the
state and how they choose each action from it (Table~\ref{tab:kinds}): a VLA
encodes the state implicitly in its weights, and an Agent Harness
(Agent-as-Policy or Harness VLA) stores it as text in a model's context.

\begin{table}[t]
\centering
\caption{\textbf{Four ways to build a robot policy with foundation models},
compared on the properties of the three advantages in Section~\ref{sec:compare}.
\cmark{} yes, \pmark{} in part, \xmark{} no. In the rest of the paper, we refer
to Agent-as-Policy and Harness VLA together as an Agent Harness.}
\label{tab:kinds}
\footnotesize
\setlength{\tabcolsep}{3.5pt}
\setlength{\aboverulesep}{0pt}\setlength{\belowrulesep}{0pt}
\renewcommand{\arraystretch}{1.38}
\newcommand{\kk}[1]{\makebox[1.25em][l]{#1}}
\newcommand{\grp}[1]{\multicolumn{5}{l}{\cellcolor{etmink!12}\textbf{#1}}}
\begin{tabular}{l llll}
\toprule
 & \textbf{Agent-as-Policy} & \textbf{Harness VLA} & \textbf{VLA / WAM} & \cellcolor{figsage!70}\textbf{COAP (ours)} \\
\midrule
\grp{Explicit State} \\
Inspectable & \kk{\pmark}reasoning text & \kk{\pmark}memory text & \kk{\xmark}black box & \cellcolor{figsage!45}\kk{\cmark}state and rules \\
Persistent & \kk{\pmark}model context & \kk{\pmark}text memory & \kk{\xmark}recent frames & \cellcolor{figsage!45}\kk{\cmark}code variables \\
Accurate & \kk{\pmark}pixels, some tools & \kk{\pmark}pixels, some tools & \kk{\xmark}image pixels & \cellcolor{figsage!45}\kk{\cmark}measured values \\
\midrule
\grp{Execution} \\
Controllable & \kk{\pmark}steered by prompt & \kk{\pmark}steered by prompt & \kk{\xmark}weights decide & \cellcolor{figsage!45}\kk{\cmark}code decides \\
Recoverable & \kk{\pmark}retries via LLM & \kk{\pmark}retries via memory & \kk{\xmark}no state to return & \cellcolor{figsage!45}\kk{\cmark}backtrack, re-attempt \\
Efficient & \kk{\xmark}LLM per decision & \kk{\xmark}LLM + VLA & \kk{\pmark}network per step & \cellcolor{figsage!45}\kk{\cmark}CPU, no model \\
\midrule
\grp{Extensibility} \\
Reusable & \kk{\pmark}base model only & \kk{\xmark}memory per task & \kk{\pmark}fine-tune per task & \cellcolor{figsage!45}\kk{\cmark}shared library \\
Fast to evolve & \kk{\pmark}prompt edits & \kk{\pmark}slow, linear gains & \kk{\xmark}data, retraining & \cellcolor{figsage!45}\kk{\cmark}reuse, inherit library \\
Cumulative & \kk{\pmark}prompts affect all & \kk{\pmark}task memory only & \kk{\xmark}tasks overwrite & \cellcolor{figsage!45}\kk{\cmark}performance accumulates \\
\bottomrule
\end{tabular}
\end{table}

\paragraph{Vision-language-action models.}
RT-2, OpenVLA, $\pi_0$, and $\pi_{0.5}$ learn language-conditioned visuomotor
policies end to end \cite{brohan2023rt2,kim2024openvla,black2025pi0,intelligence2025pi05},
and recent foundation models scale data and training further
\cite{wu2026pragmatic,kim2026rldx}. A VLA cannot be inspected
or edited directly, so each improvement requires new data and a new training
run, and LIBERO-PRO shows that benchmark success can reflect memorized
training configurations \cite{zhou2025liberopro}. In comparison, we keep the state in code
variables, so we can \textbf{backtrack from a failure to the
variable that caused it} and fix the code logic that derived it. Each accepted fix
is merged into the code without breaking existing logic, so \textbf{improvements accumulate}, while training a
VLA on a new task can degrade its performance on earlier tasks
\cite{zhu2026vla}.

\paragraph{VLM (System-2) for robot control.}
ReAct interleaves reasoning with actions \cite{yao2023react}, and Toolformer
teaches a model to call tools \cite{schick2023toolformer}. In robotics,
SayCan grounds language-model plans in learned affordances
\cite{ahn2022saycan}, and Inner Monologue feeds environment feedback back to
the planner \cite{huang2022inner}. Recent harnesses let a VLM agent call
analytic or learned primitives \cite{chen2026showharness,zhang2026harness},
and GPT-6 Astra~\cite{openai2026astra} has been evaluated directly as an
embodied policy \cite{su2026astra,zhang2026astra}. In these systems, the model is queried repeatedly during execution to output actions, so there is randomness at each call, even if the prompt is the same. Moreover, every episode is of high cost due to extensive
model calls \cite{openai2026pricing}. In our approach, models are used only
offline: coding agents write the code, and the robot runs it without a
model during online execution. The code computes each action from the state
through explicit rules, so \textbf{the robot's behaviour is reproducible},
and \textbf{each decision is cheap} because model calls are not required.

\paragraph{Code-as-Policy.}
Code-based policies differ in when the program is written and in the source
of the state used (Figure~\ref{fig:code_policies}). Code-as-Policies and
ProgPrompt prompt a language model to write code that calls perception and
control APIs \cite{liang2023code,singh2023progprompt}, and VoxPoser writes code that
composes spatial value maps \cite{huang2023voxposer}. Later works add
execution feedback, repeated generation, and skill discovery
\cite{fu2026capx,naouali2026vlcp,lu2026aspire}. PyRUA-Lean has a model
write code cell by cell as each episode runs \cite{si2026pyrua}, and
Skill2Real has a model compose a new program from frozen skill memories at
each decision stage \cite{he2026skill2real}. In these systems, a model writes
the program at runtime, and the state is provided by learned perception
models:
object detectors in Code as Policies, and SAM in PyRUA-Lean and Skill2Real. In RHO and
EmbodiedSWE, coding agents write one program offline, but its state still
comes from outside the code: RHO likewise obtains it from SAM and Molmo
\cite{elmaaroufi2026rho,carion2025sam3}, and the solvers of EmbodiedSWE read
the simulator's oracle values \cite{shen2026embodiedswe}. These perception
models are black boxes, and their uncertainty are not controlled by the code. In contrast, our code tracks the state
itself, so \textbf{the state is programmable}: when a measurement is wrong,
coding agents fix the measurement code, and \textbf{the state becomes more
accurate over time}.

\paragraph{Code as an interface for agents.}
Outside robotics, code has become the medium through which agents act and
improve. CodeAct uses executable code as the action space of an agent
\cite{wang2024codeact}, and exposing tools as a code API lets a model compose
them with fewer tokens \cite{anthropic2025mcp,varda2025codemode}. Voyager grows
a library of code skills through interaction \cite{wang2023voyager}, and
software agents solve long-horizon repository tasks through tests and diffs
\cite{yang2024sweagent,openai2025codexcli}. In these settings, the environment
provides the state explicitly, as files, test results, or game API values. A
robot, in contrast, must measure its state from sensors, and our code performs
this measurement itself. In robotics, RoboRSI runs RSI as a loop of manager,
planner, engineer, and reviewer agents with a no-regression gate, but a VLM
still makes the decisions during execution \cite{noematrix2026roborsi}. A
change that passes the gate only steers this black box, whose uncertainty and
errors cannot be fully controlled, so the tested behaviour may not be the
behaviour at runtime. Our policy is code with explicit state, so the tested
behaviour is the behaviour at runtime, and coding agents can improve it the way
software engineers improve a program: \textbf{every change is tested offline},
and \textbf{only changes that improve results without regressions on other
tasks are merged}.

\paragraph{Benchmarks.}
LIBERO measures knowledge transfer in lifelong robot learning
\cite{liu2023libero}, RoboTwin~2.0 generates domain-randomized bimanual tasks
\cite{chen2025robotwin2}, and RoboCasa365 scales household simulation
\cite{nasiriany2026robocasa365}. RoboDojo spans memory, precision, open
instructions, generalization, and long-horizon manipulation in one protocol
\cite{chen2026robodojo}. Its memory and precision tasks directly test how
well a policy retains and measures state, and the shared protocol lets us compare
VLAs, world models, agents, and code on the same 42 tasks.

\section{Embodied Turing Machine: Code-Only-as-Policy}
\label{sec:method}

\begin{figure}[t]
\centering
\includegraphics[width=\linewidth]{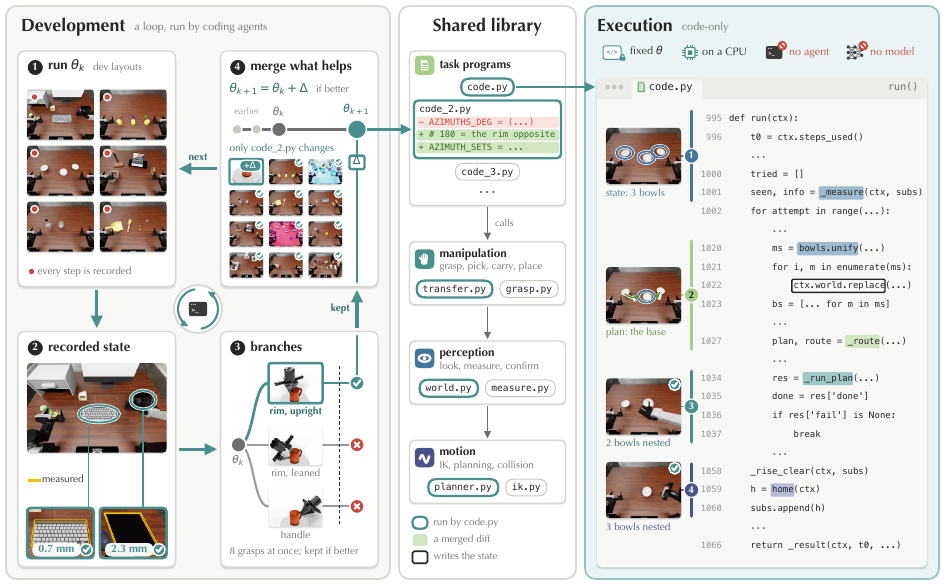}
\caption{\textbf{Code-Only-as-Policy.} \emph{Development} (left): coding
agents improve the code offline in a loop. They run the current code
$\theta_k$ on development layouts, read the state, and run
alternative diffs on branches in parallel.
\emph{Shared library} (middle): task programs call the library from the shared code. \emph{Execution} (right): at runtime the robot runs one
fixed task program across episode, and an agent and a VLA are not required.}
\label{fig:method}
\end{figure}

Code-Only-as-Policy implements the Embodied Turing Machine
(\S\ref{sec:introduction}) as a shared code library, in which code variables
hold the state and the source code constitutes the rules. Beyond storing the
state, the perception code measures the environment from
camera images, and the code updates these measurements as the robot acts. The
code is frozen at runtime and improved offline by coding agents
(Figure~\ref{fig:method}). The development does not involve test set of a benchmark. 

\subsection{Reading and writing the state}

At step $t$, RoboDojo provides an observation $o_t$ consisting of an
instruction, three RGB images (head and two wrists), joint angles,
end-effector poses, and the gripper command echo \cite{chen2026robodojo}. The
code maintains an explicit state $x_t$ with three parts, each of which has the
same form in every task. (i) The \textbf{environment state} records the pose,
size, and category of each object and the spatial relations between objects,
such as one object resting on another; each record also stores the error,
source, and time of its measurement. (ii) The \textbf{robot state} records the
joint angles, end-effector pose, and gripper opening of each arm, together
with the poses each arm can reach. These two parts represent the tape of the
Embodied Turing Machine. (iii) The \textbf{task state} records the current
stage, the arm assigned to each object, retry counts, and the information the
task must retain, such as the color under each cup. It is not part of the
tape and plays the role of the internal state of a Turing machine.

Each part is updated from a different source. The robot state is obtained
from proprioception at every step. The environment state is measured by
perception layer library. The task code updates the task state during execution.

The source code $\theta$ constitutes the rules. It defines the transition
\begin{equation}
 (x_{t+1},\,a_t)=F_\theta(x_t,\,o_t),
 \label{eq:transition}
\end{equation}
where $a_t$ is the next trajectory segment. The code $\theta$ is frozen at
runtime, and only $x_t$ changes. A new episode layout therefore changes only the
state, so one program runs in every episode layout, including the held-out test
layouts (\S\ref{sec:setup}).

Code Example~\ref{lst:core} shows the core of a task program that covers three cubes with
cups and then uncovers them in a fixed color order; it uses all three parts
of the state. \texttt{reg} holds the environment state: the measured
records of three cubes and three cups, which the code pairs into cube-cup
slots ordered from left to right. \texttt{memory} and \texttt{stage} hold the
task state: the color under each cup and the active subgoal.
The move step, shown in teal color, uses the robot state to select an arm and
compute each motion. After the third move, the colors are no longer visible, so the code
retrieves them from \texttt{memory} to determine the target of each uncover
move.

\begin{figure}[t]
\codelabel{lst:core}
\begin{codewindow}{code\_4.py}
(*@\lnum{131}@*)def run(ctx):
    (*@\skipped@*)
(*@\lnum{133}@*)    snap = ctx.snap(('head',))
(*@\lnum{134}@*)    cubes = (*@\hlP{blocks.measure\_cubes}@*)(ctx, snap, edge=CUBE_EDGE, colours=COLORS, n=3)
(*@\lnum{135}@*)    cups = (*@\hlP{containers.measure\_cups}@*)(ctx, snap, inverted=True, n=3)
    (*@\skipped@*)
(*@\lnum{141}@*)    reg = (*@\hlS{ctx.register}@*)(list(cubes) + list(cups))
(*@\lnum{142}@*)    slots = blocks.slots_by_x(reg[:3], reg[3:])
(*@\lnum{143}@*)    order = [('cover', i) for i in range(3)] + [('uncover', blocks.slot_of(slots, c)) for c in COLORS]
(*@\lnum{144}@*)    (*@\hlS{memory}@*) = {s.colour: s.index for s in slots}
    (*@\skipped@*)
(*@\lnum{146}@*)    for k, (op, i) in enumerate(order):
(*@\lnum{147}@*)        (*@\hlS{stage}@*) = SPEC.stages[k]
        (*@\skipped@*)
(*@\lnum{153}@*)        cup, _src, dst = containers.move_for(slots, i, op)
(*@\lnum{154}@*)        outs, placed, verified = (*@\hlM{\_move\_cup}@*)(ctx, cup, dst, stage, HOME_EST_STEPS)
\end{codewindow}
\codecaption{\textbf{Code Example~\thecodelisting.} The core of a task program. As in
Figure~\ref{fig:method}, perception calls are blue, the manipulation step is
teal, and writes to the state are outlined.}
\end{figure}

\subsection{Measuring the environment state}

Perception is also ordinary code, built on NumPy and OpenCV, so the state is
programmable: every measurement can be inspected and revised like any other
rule. In each frame, the robot's mesh is projected into the image using the
measured joint angles, and the pixels it covers are masked out. Pixels that
differ from the current table color form candidate regions, and each
connected region whose lower edge rests on the table becomes an object
candidate. For a boundary pixel with camera centre $c$ and ray direction $d$,
intersecting the ray with the table plane $z=z_0$ yields its position on the
table:
\begin{equation}
 p=c+\lambda d,\qquad \lambda=(z_0-c_z)/d_z .
 \label{eq:plane}
\end{equation}
A perception for each kind of object then places the known 3D shape of the
candidate near $p$, searches over its position and yaw, and keeps the fit with
the largest overlap with the observed region. The camera models, robot meshes,
object shapes, and table height used in these steps are published with the
benchmark and do not change between episodes, so they are constants in the
code, part of the rules $\theta$. What changes between episodes, such as which
objects are present and where they are, is state, which the code measures from
the images in every episode.

Each matched measurement yields an object record with category, position, yaw, size, and
confidence. Records persist in an episode-level \texttt{WorldState}: an
object occluded by the arm keeps its last measured pose. Tasks that require
millimetre accuracy re-measure the object with a wrist camera before the
final approach.

\subsection{Sharing one library across tasks that read the same state}

Two design choices make the library reusable across tasks. (i)~Every task
operates on state of the same form (\S3.1), so a module written for one task
can be reused by other tasks. (ii)~The code is layered, with dependencies
pointing downward (Figure~\ref{fig:method}, middle): task programs call
object-specific code, which calls manipulation code such as grasp, pick,
place, push, and pour, which in turn calls perception and motion code. Task
code specifies \emph{what} to do: the order of subgoals, task-specific
decisions, and recovery conditions. The lower layers determine \emph{how}. For
example, a single \texttt{transfer} call expands into grasp-candidate
generation, inverse-kinematics and collision filtering, a pre-grasp
re-measurement, approach, closure, a short lift with a hold check, carry,
placement, release, and a final verification, each of which returns an
outcome with measured evidence. As a result, 83\% of the lines are shared across tasks (\S\ref{sec:cmp-ext}).

A new task extends the library in one of three ways (Code Example~\ref{lst:extend}). It can
\emph{compose} existing modules, such as \texttt{transfer}. It can
\emph{inherit} an existing task and override a single decision: a task that
sorts objects into the baskets named in its instruction reuses the entire flow
of an existing sorting task and replaces only the rule that assigns objects to
baskets. Or it can add a
\emph{default-off parameter} to a shared function: \texttt{lift\_dir} allows a
grasped object to be lifted diagonally, which one task requires near a
basket wall, while the default preserves the existing
vertical lift.

\begin{figure}[t]
\codelabel{lst:extend}
\begin{codewindow}{grasp.py}
(*@\lnum{122}@*)LIFT_UP = (0.0, 0.0, 1.0)     # default: the established vertical lift
(*@\skipped@*)
(*@\lnum{312}@*)def execute_grasp(c, ctx, attempt=1, ..., (*@\hlT{lift\_dir=LIFT\_UP}@*),
(*@\lnum{313}@*)                  lift_speed=None, lift_accel=None, band_at_close=None):
\end{codewindow}
\begin{codewindow}{code\_5.py}
(*@\lnum{46}@*)class Rules((*@\hlT{CO.Rules}@*)):              # inherit an existing task's flow
    (*@\skipped@*)
(*@\lnum{57}@*)    def (*@\hlT{first}@*)(self, ctx, items, info):    # override one decision
(*@\lnum{58}@*)        sort_language.label(items, self.cats, info.get('frame'))
        (*@\skipped@*)
(*@\lnum{73}@*)def run(ctx):
    (*@\skipped@*)
(*@\lnum{82}@*)    return (*@\hlT{CO.run}@*)(ctx, Rules(cats))
\end{codewindow}
\codecaption{\textbf{Code Example~\thecodelisting.} A default-off parameter of a shared function
(top) and a task that inherits another task and overrides one decision
(bottom); the new code is highlighted.}
\end{figure}

Default-off parameters provide a compatibility guarantee. Let
$B_i(\theta,o)$ be the execution trace of task $i$ under code $\theta$ and
observations $o$. For every task $i$ that does not pass the new argument, an
extension $\delta$ must satisfy
\begin{equation}
 B_i(\theta+\delta,\,o)=B_i(\theta,\,o)\qquad \forall\, o\in\mathcal{O}_{\rm test},
 \label{eq:compatibility}
\end{equation}
where $\mathcal{O}_{\rm test}$ denotes the fixtures exercised by the unit
tests. Existing tasks therefore retain their behaviour by construction, and
only the adopting task changes.

\subsection{Improving the code by backtracking failures to the state}

Because the code records its state at every step of an episode, coding
agents can improve the code offline from these records. Improvement proceeds
in rounds and is carried out by Claude Opus 5.5 coding agents, which wrote all
code in the library and in every task, ran the tests and evaluations, and
debugged their own code autonomously. Let $\theta_k$ be the accepted code
after round $k$. Each agent works in its own git worktree, inspects the
recorded state of failed development runs to locate the variable or rule at
fault, and proposes a diff $\Delta$. The resulting branches form a search
frontier of runnable code:
\begin{equation}
 \mathcal{C}_k=\{\theta_k+\Delta_{k,1},\,\ldots,\,\theta_k+\Delta_{k,m}\}.
 \label{eq:frontier}
\end{equation}
Because each branch is complete, runnable code, branches can be evaluated
concurrently under identical conditions. Before simulation, each branch passes
two CPU checks: the task code is executed against a numeric scene to confirm
that every stage reaches its target state, and the robot and object meshes
are checked for inverse-kinematics solutions and collisions at every
waypoint. Branches that pass are evaluated on the validation episodes
(\S\ref{sec:setup}) in parallel.

Debugging and selection use separate data. Agents debug on a development set
they define themselves, where they watch videos and compare the measured
state against simulator ground truth; selection compares success counts on
the validation split. Let $J_q(\theta;V_q)$ be the success count of task $q$
on these layouts and $\epsilon_q$ a margin above rerun noise. A branch is
accepted if
\begin{equation}
 J_q(\theta_k+\Delta;\,V_q)\;\geq\;J_q(\theta_k;\,V_q)+\epsilon_q ,
 \label{eq:accept}
\end{equation}
in which case it becomes $\theta_{k+1}$ and the base of all subsequent
branches. Otherwise, the current code remains deployed and the diff is not
merged. The compatibility condition (Equation~\ref{eq:compatibility})
protects the tasks that a diff does not target, and the acceptance rule
(Equation~\ref{eq:accept}) ensures that the targeted task improves, so each
accepted step is non-decreasing and improvements accumulate.

\section{Experiments}
\label{sec:experiments}
\begin{table}[t]
\centering
\caption{\textbf{Results on RoboDojo.} SR is the success rate (\%), the share of episodes in which the whole task succeeds; every column except Score reports SR. Score is the benchmark's progress score.}
\label{tab:main}
\footnotesize
\setlength{\tabcolsep}{5pt}
\renewcommand{\arraystretch}{1.08}
\begin{tabularx}{\linewidth}{@{}>{\raggedright\arraybackslash}X*{7}{c}@{}}
\toprule
Method & Memory & Precision & Open & Generalization & Long-Horizon & \multicolumn{2}{c}{Overall} \\
\cmidrule(l){7-8}
 & (6) & (8) & (8) & (12) & (8) & SR & Score \\
\midrule
\multicolumn{8}{@{}l}{\blocktitle{Agent Harness}} \\
PhysicalRSI~\cite{physicalrsi2026} {\scriptsize\color{etmink!70}(agent + VLA)} & 46.6 & 32.5 & 24.9 & 15.6 & 37.3 & \underline{31.38} & \underline{36.27} \\
GPT-6 Astra~\cite{zhang2026astra} & 38.7 & 4.0 & \underline{31.0} & \underline{30.5} & 8.2 & 22.48 & 28.97 \\
\midrule
\multicolumn{8}{@{}l}{\blocktitle{World Action Model (WAM)}} \\
Awomo-0.5~\cite{awomo2026} & 41.2 & \underline{33.5} & 22.9 & 23.8 & 26.8 & 29.64 & 35.34 \\
VPP2-Preview~\cite{vpp22026} & \underline{50.3} & 25.6 & 5.4 & 24.6 & 22.2 & 25.62 & 31.40 \\
Liber-0 Preview~\cite{liber2026preview} & 37.3 & 33.2 & 5.3 & 18.6 & 33.2 & 25.52 & 30.74 \\
Liber-0 Lite~\cite{liber2026lite} & 35.4 & 31.1 & 2.6 & 19.1 & 32.9 & 24.23 & 29.24 \\
InternW0-$\Delta$~\cite{miao2026internw0} & 34.0 & 23.2 & 10.2 & 22.8 & 29.3 & 23.91 & 30.77 \\
ME-Brain-1.0~\cite{he2026mebrain} & 22.8 & 15.9 & 5.3 & 15.3 & 20.6 & 15.99 & 21.67 \\
OpenWAM-$\alpha$~\cite{wang2026openwam} & 9.1 & 9.2 & 1.1 & 14.8 & 25.3 & 11.92 & 17.18 \\
\midrule
\multicolumn{8}{@{}l}{\blocktitle{VLA Model}} \\
Simate-beta~\cite{simate2026} & 33.0 & 26.9 & 8.5 & 27.9 & \underline{43.4} & 27.96 & 33.95 \\
Rex-M1 Preview~\cite{rexm12026} & 48.2 & 23.8 & 6.1 & 26.1 & 33.0 & 27.44 & 33.79 \\
DM0.5~\cite{dm052026} & 47.4 & 16.8 & 2.1 & 10.9 & 19.5 & 19.34 & 24.90 \\
GalaxeaVLA (G0.5)~\cite{galaxea2026g05} & 7.3 & 20.4 & 1.6 & 12.8 & 32.2 & 14.88 & 20.23 \\
Xiaomi-Robotics-1~\cite{xiaomi2026robotics1} & 6.6 & 18.8 & 3.6 & 17.0 & 23.7 & 13.93 & 20.07 \\
Meituan-Robotics-0~\cite{meituan2026robotics0} & 8.9 & 7.8 & 4.2 & 8.2 & 18.6 & 9.53 & 14.95 \\
SimpleMemVLA~\cite{yin2026simplememvla} & 33.2 & 2.9 & 0.8 & 3.9 & 5.5 & 9.27 & 12.58 \\
Hy-Embodied-0.5-VLA~\cite{zhang2026hyembodied} & 12.1 & 8.0 & 0.6 & 8.4 & 14.9 & 8.80 & 13.07 \\
KinRT~\cite{yang2026kinrt} & 3.6 & 9.9 & 3.8 & 8.6 & 18.1 & 8.80 & 13.02 \\
Spatial Forcing~\cite{li2025spatialforcing} & 4.1 & 10.6 & 1.6 & 9.3 & 14.6 & 8.04 & 12.38 \\
VLAct~\cite{yang2026vlact} & 0.6 & 15.2 & 2.2 & 6.3 & 13.7 & 7.58 & 10.65 \\
StarVLA-PI\_v3~\cite{starvla2026codebase} & 4.0 & 12.5 & 2.0 & 8.1 & 11.0 & 7.51 & 10.81 \\
InternVLA-A1.5~\cite{ma2026internvla} & 3.6 & 10.2 & 1.4 & 6.8 & 13.8 & 7.14 & 11.15 \\
$\pi_{0.5}$~\cite{intelligence2025pi05} & 4.7 & 5.5 & 1.7 & 8.2 & 14.7 & 6.93 & 11.44 \\
\midrule
\multicolumn{8}{@{}l}{\blocktitle[ropediadeepgreen]{Code-Only-as-Policy (COAP)}} \\
\rowcolor{coapbg}
\textbf{COAP (ours)} & \textbf{89.9} & \textbf{75.1} & \textbf{64.1} & \textbf{65.7} & \textbf{56.4} & \textbf{70.24} & \textbf{75.45} \\
\rowcolor{coapbg}
\emph{vs.\ best} & \textcolor{dahaccent}{+39.6} & \textcolor{dahaccent}{+41.6} & \textcolor{dahaccent}{+33.1} & \textcolor{dahaccent}{+35.2} & \textcolor{dahaccent}{+13.0} & \textcolor{dahaccent}{+38.86} & \textcolor{dahaccent}{+39.18} \\
\bottomrule
\end{tabularx}
\end{table}

This section answers three questions: (1) How well does Code-Only-as-Policy
(COAP) perform on a hard benchmark (\S\ref{sec:setup}--\S\ref{sec:main})?
(2) What does COAP offer that a VLA or an Agent Harness lacks
(\S\ref{sec:compare})? (3) Why does COAP suit RSI (\S\ref{sec:dev})?

\subsection{Setup}
\label{sec:setup}

\paragraph{Benchmark and protocol.}
RoboDojo~\cite{chen2026robodojo} contains 42 bimanual tasks that span five
dimensions: Memory, Precision, Open instruction following, Generalization, and
Long-Horizon manipulation. Each task is evaluated with three seeds of 50
layouts each. In the Generalization tasks, half of the layouts are randomized,
with new object instances, a new room and table appearance, and up to 20
distractors. All reported numbers are produced by the official evaluation
client, judges, and summary script.

\paragraph{Inputs.}
The policy receives only three RGB camera streams, proprioception (joint
angles, end-effector poses, and the echo of the gripper command), and the
language instruction. COAP uses no depth, no segmentation, and no model service
at test time, and runs efficiently and cheaply.

\paragraph{Baselines.}
We compare with the official RoboDojo leaderboard~\cite{chen2026robodojo}.
Table~\ref{tab:main} lists its 23 entries at or above $\pi_{0.5}$, grouped
into three categories. Agent Harness methods include
\mbox{PhysicalRSI}~\cite{physicalrsi2026} and \mbox{GPT-6} Astra~\cite{zhang2026astra}.
World Action Models (WAMs) include \mbox{Awomo-0.5}~\cite{awomo2026},
VPP2-Preview~\cite{vpp22026}, \mbox{Liber-0} Preview~\cite{liber2026preview},
\mbox{Liber-0} Lite~\cite{liber2026lite},
\mbox{InternW0-$\Delta$}~\cite{miao2026internw0},
ME-\mbox{Brain-1.0}~\cite{he2026mebrain}, and
\mbox{OpenWAM-$\alpha$}~\cite{wang2026openwam}. VLA models include
Simate-beta~\cite{simate2026}, Rex-M1 Preview~\cite{rexm12026},
DM0.5~\cite{dm052026}, \mbox{GalaxeaVLA} (G0.5)~\cite{galaxea2026g05},
Xiaomi-\mbox{Robotics-1}~\cite{xiaomi2026robotics1},
Meituan-\mbox{Robotics-0}~\cite{meituan2026robotics0},
SimpleMemVLA~\cite{yin2026simplememvla},
Hy-\mbox{Embodied-0.5}-VLA~\cite{zhang2026hyembodied},
KinRT~\cite{yang2026kinrt}, Spatial Forcing~\cite{li2025spatialforcing}, VLAct~\cite{yang2026vlact}
and StarVLA-PI\_v3~\cite{starvla2026codebase}, InternVLA-A1.5~\cite{ma2026internvla},
and $\pi_{0.5}$~\cite{intelligence2025pi05}.

\subsection{Main results}
\label{sec:main}

Table~\ref{tab:main} compares Code-Only-as-Policy (COAP) with the leaderboard.
COAP achieves a success rate of 70.24\%, 38.9 percentage points above the current
state-of-the-art method (31.38\%), and more than 40 points above
the strongest World Action Model (29.64\%), and the strongest VLA (27.96\%). COAP also has the highest success rate in each of the
five dimensions. The margins are largest on Memory (89.9\%, 39.6 points above
the best baseline) and Precision (75.1\%, 41.6 points), the two dimensions in
which the next action depends on a stored value or a precise measurement. COAP
reaches 65.7\% on Generalization (77.3\% on standard and 54.1\% on randomized
layouts) and 64.1\% on Open. Its margin is smallest on
Long-Horizon tasks.

\subsection{Advantages of Code-Only-as-Policy}
\label{sec:compare}

\begin{figure}[!h]
\centering
\includegraphics[width=0.72\linewidth]{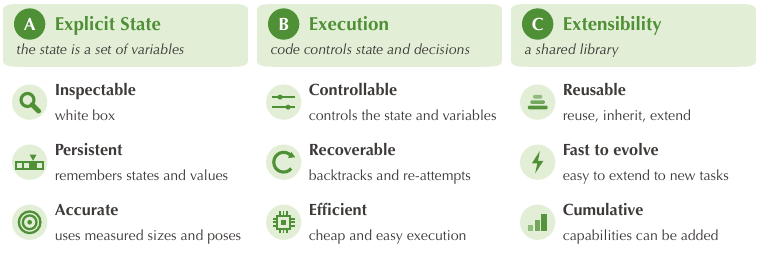}
\caption{\textbf{Three advantages of COAP}, each with three properties.}
\label{fig:adv}
\end{figure}

We compare Code-Only-as-Policy (COAP) with a VLA and an Agent Harness on three
advantages, summarized in Figure~\ref{fig:adv}: \textbf{Explicit State}
(\S\ref{sec:cmp-state}), \textbf{Execution} (\S\ref{sec:cmp-exec}), and
\textbf{Extensibility} (\S\ref{sec:cmp-ext}). For each baseline, we use the
strongest leaderboard entry of its category (Table~\ref{tab:main}). Where the
benchmark provides no measurement, we use numbers reported in prior
work~\cite{black2025pi0,intelligence2025pi05,kim2024openvla,zhu2026vla}.

\subsubsection{Explicit State}
\label{sec:cmp-state}

\finding{1}{Code-Only-as-Policy stores the robot, environment, and task state as
explicit variables and computes actions from them. In comparison, an Agent
Harness keeps the state in text memory and its model context, and a VLA keeps
it implicitly in its weights.}

\begin{figure}[!b]
\centering
\includegraphics[width=\linewidth]{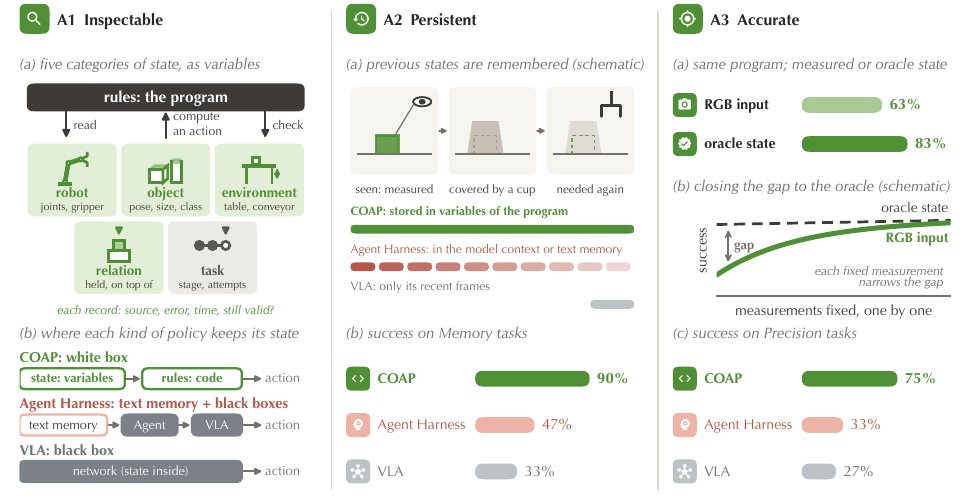}
\caption{\textbf{Explicit State.} COAP keeps five categories of state as
variables (left), remembers states and values (middle), and succeeds more often
with more accurate measurements (right).}
\label{fig:A}
\end{figure}

\advpoint{A1}{Inspectable.} Code-Only-as-Policy is a white box. What the
program perceives about the robot and the environment becomes explicit state in
code: variables, data structures, and the parameters of functions. As shown in
Figure~\ref{fig:A} (left), the state has five categories: robot, object,
environment, relation, and task. The coding agent can therefore write the
decisions in advance as code over these values, so no agent is needed during
online execution. In comparison, in a VLA, both the state and the reasoning are
implicit in the weights. An Agent Harness keeps the task memory, but its
decisions still come from model inference: an agent or a VLA still does the
reasoning, so some states stay hidden in this model-in-the-loop system, which
cannot be fully inspected (even the same memory can lead to different actions
at different calls).

\advpoint{A2}{Persistent.} COAP stores each measured value in a variable of the
program, and the value persists throughout the execution. For example, when a
static object is covered, its position and pose are still tracked in code, as
shown in Figure~\ref{fig:A} (middle). Decisions that depend on memory, such as
following a specific order, are also written explicitly in code. In comparison,
a VLA sees only its recent frames, and an Agent Harness keeps what it has seen
only in its model context or text memory. Memory-related tasks can therefore be
solved easily by code, since the state is tracked and stored persistently in
the program: on the Memory tasks of RoboDojo, COAP reaches 90\%, against 47\%
for the Agent Harness and 33\% for the VLA.

\begin{figure}[!b]
\centering
\includegraphics[width=\linewidth]{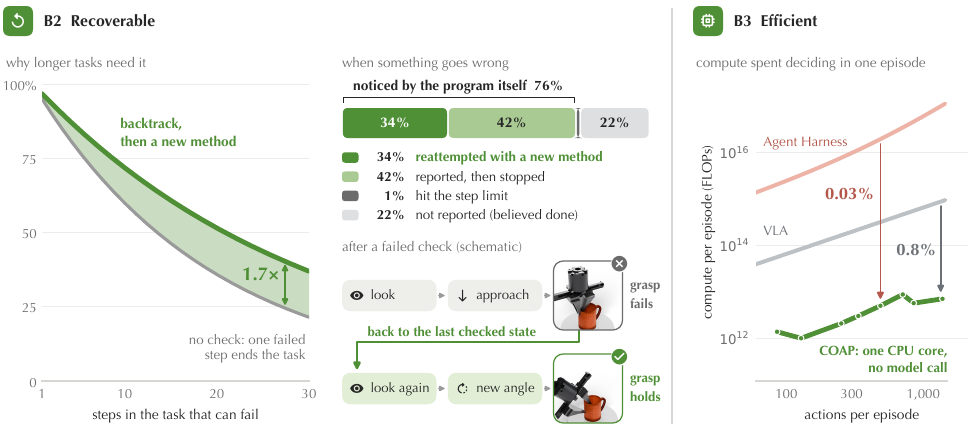}
\caption{\textbf{Execution.} Backtracking keeps failures from compounding over
long tasks (left, middle), and code runs on a CPU with a small fraction of a
model's compute (right; COAP's compute as a share of each model's, from
reported model sizes).}
\label{fig:B}
\end{figure}

\advpoint{A3}{Accurate.} COAP acts on measured quantities, because the
measurements can be estimated offline without accessing the simulator's oracle
values. The cameras, the robot's 3D meshes, and the sizes and appearance of the
objects are estimated before deployment, so at test time the program only
locates each object, its position, size, and pose, and sets the grasp width and
the clearance from them. During development, the oracle values help improve the
perception functions, so the success rate on measured state can approach that
on oracle values: as shown in Figure~\ref{fig:A} (right), the same program
succeeds in 63\% of episodes on measured state and in 83\% on oracle state, and
each measurement that is fixed narrows this gap. In comparison, a VLA and an
Agent Harness act mainly on image pixels and never make measurements explicit.
COAP therefore has a higher upper bound on tasks that need precision: on the
Precision tasks of RoboDojo, it reaches 75\%, against 33\% for the Agent Harness
and 27\% for the VLA.

\subsubsection{Execution}
\label{sec:cmp-exec}

\finding{2}{Code-Only-as-Policy controls the state and the decision making.
Because the state is tracked, failures can be recovered by backtracking and re-attempting through the robust code design. Online execution is efficient and cheap since it is model-free.}

\advpoint{B1}{Controllable.} COAP has no model in the online execution, so code
controls the state, its variables, and the decision making. For example, when a
bowl held by its rim slipped during fast carries, one line of code limited the
carry speed of rim grasps, and the bowl no longer dropped. In comparison, a VLA,
or a model in the loop of an Agent Harness, is a black box: some of its
uncertainty and errors cannot be fully controlled and accumulate over the
execution. More training trajectories for a VLA, or more text memory for an
Agent Harness, still leave its decisions uncertain.

\advpoint{B2}{Recoverable.} Because the state is tracked and controlled, the
program knows where the task stands after each step, so it can backtrack and
re-attempt. As shown in Figure~\ref{fig:B} (middle, bottom), when an initial
grasp fails, the program backtracks, measures the object again, and re-attempts
with a new grasp method, such as a new grasp angle. In episodes where something
went wrong, the program noticed the problem itself in 76\% and re-attempted
with a new method in 34\% (Figure~\ref{fig:B}, middle). Recovery matters more as
tasks grow longer, since without it one failed step ends the task
(Figure~\ref{fig:B}, left). In comparison, a VLA has no tracked state to return
to.

\advpoint{B3}{Efficient.} COAP uses a model only offline, when the coding agent
writes the code, and the code is shared by every episode of a task. A control
step takes a median of 0.3\,ms on a CPU, and an episode uses 0.8--2.4\% of the
compute of a VLA and less than 0.1\% of that of an Agent Harness
(Figure~\ref{fig:B}, right). In comparison, an Agent Harness, especially
GPT-as-policy, is much slower and more expensive, and it reasons anew in every
episode instead of sharing its decisions across the episodes of a task.

\subsubsection{Extensibility}
\label{sec:cmp-ext}

\finding{3}{Code-Only-as-Policy builds every new task on a shared library by
reusing, inheriting, or extending its code. Each new task is therefore fast to
develop, and capabilities accumulate without harming the performance on old tasks.}

\begin{figure}[!t]
\centering
\includegraphics[width=\linewidth]{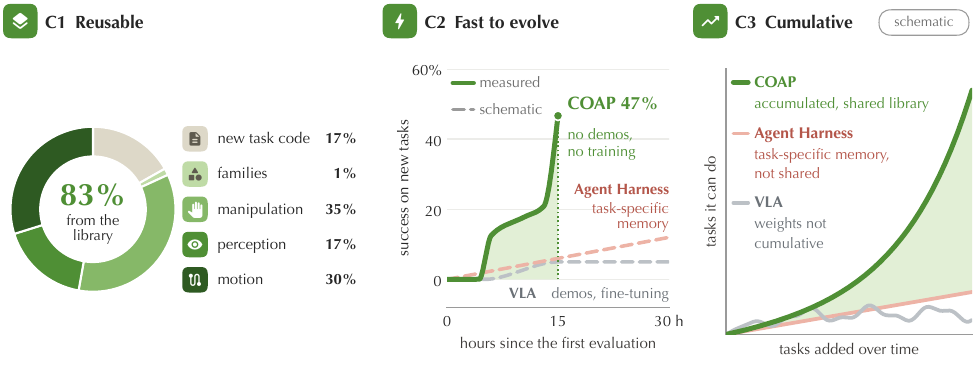}
\caption{\textbf{Extensibility.} New tasks run mostly library code (left),
are developed from scratch within hours without data or training (middle;
baseline lines are schematic), and add to what the policy can do (right,
schematic).}
\label{fig:C}
\end{figure}

\begin{figure}[!t]
\centering
\includegraphics[width=0.86\linewidth]{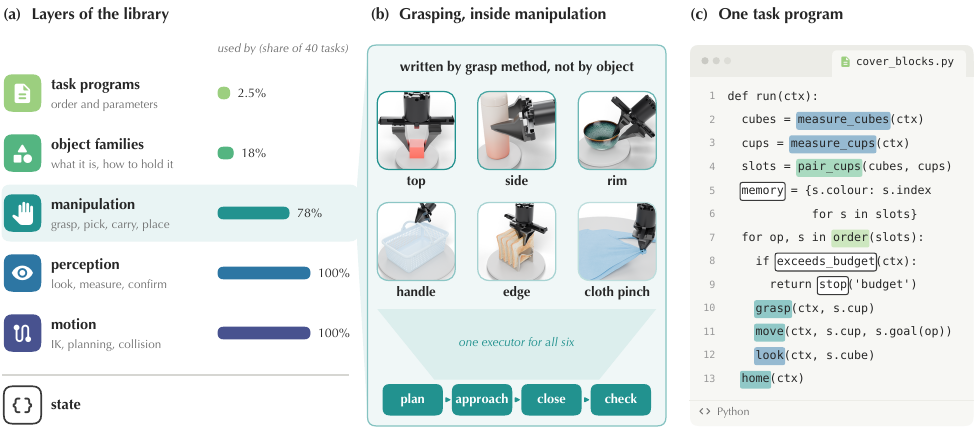}
\caption{\textbf{Code structure.} (a) The layers of the library. (b) Functions
are abstracted at a sensible level: grasping is written by grasp method, not by
object. (c) A simplified task program;
each call is colored by its layer.}
\label{fig:code}
\end{figure}

\advpoint{C1}{Reusable.} A new task \emph{reuses}, \emph{inherits}, or
\emph{extends} the shared library, so its own code holds only task-specific
logic, such as a check or a picking order. As shown in Figure~\ref{fig:C}
(left), 83\% of the code that new tasks run comes from the library. The library
can be shared because its functions are abstracted step by step at a sensible
level (Figure~\ref{fig:code}). For example, grasping is not written for each
object but by grasp method, such as from the top, by the rim, or by the handle,
so a new object can use an existing method. In comparison, a VLA needs
fine-tuning for new tasks. The memory of an Agent Harness is task-specific and
cannot be shared, and its execution is specific to each episode, so it cannot
be shared either.

\advpoint{C2}{Fast to evolve.} Since the library is shared, a new task can
reuse, inherit, or extend the existing library, so COAP evolves easily to new
tasks. Figure~\ref{fig:C} (middle) shows development on new tasks from scratch,
on tasks that had no program when the library was frozen. COAP reached an
average success of 47\% within 15 hours, without data or training. Its curve
rises steeply, because the library becomes more robust as it grows, and every
task benefits from it. The baseline lines are schematic. A VLA needs new
demonstrations and retraining; OpenVLA, for example, reports 5--15 hours of
fine-tuning per task on 8 A100 GPUs~\cite{kim2024openvla}. On the same tasks, the strongest
VLA reaches 4.7\% after training on demonstrations. An Agent Harness improves
roughly linearly and much more slowly: most of its memory is task-specific, and
only a small part, such as general insights, carries over between tasks.

\advpoint{C3}{Cumulative.} The library is designed to be refactorable: a new
task adds functions that inherit from existing ones, or options that are off by
default, instead of changing the logic of earlier tasks. Earlier tasks therefore
keep running unchanged code, and capabilities can be added without making old
tasks worse. As the code becomes more robust, each new task becomes simpler to
add, so the capabilities of COAP grow exponentially (Figure~\ref{fig:C}, right;
schematic). In comparison, an Agent Harness builds task-specific memory by
running each task and reuses little of it, so its capabilities grow linearly. A VLA must be retrained for new tasks, and
the effect on old tasks cannot be controlled: after sequential fine-tuning of
$\pi_{0.5}$ on five real-world tasks, its average score drops from 86.9 to
31.4~\cite{zhu2026vla}.

\subsection{Why Code-Only-as-Policy Suits RSI}
\label{sec:dev}

\begin{figure}[!t]
\centering
\includegraphics[width=0.92\linewidth]{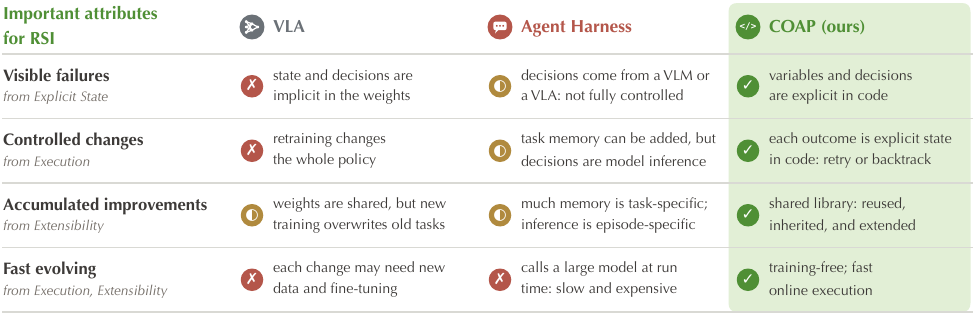}
\caption{\textbf{Why Code-Only-as-Policy is a suitable medium for RSI.} Each
row is an attribute important for RSI, which comes from an advantage in
Section~\ref{sec:compare} (check: yes; half circle: partly; cross: no).}
\label{fig:rsi}
\end{figure}

\begin{figure}[!t]
\centering
\includegraphics[width=\linewidth]{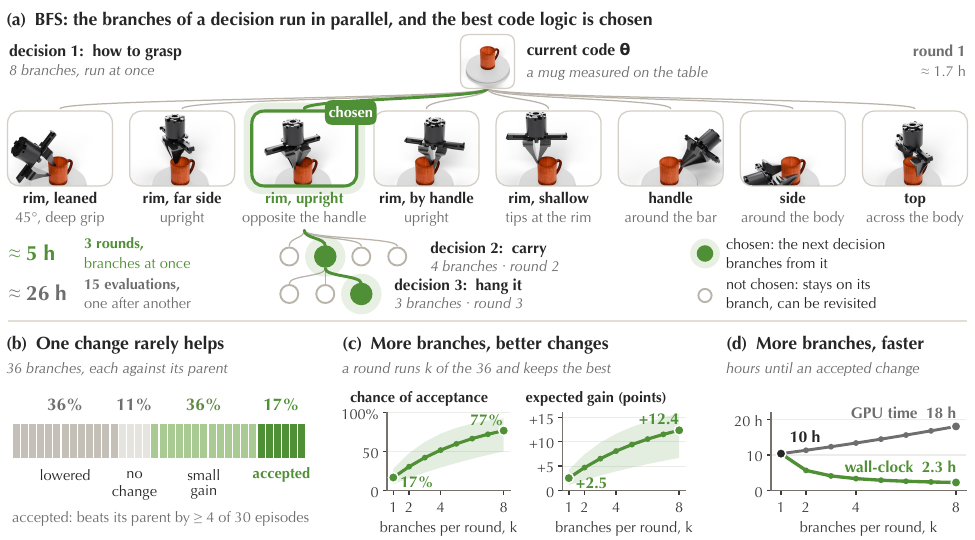}
\caption{\textbf{Parallel branching over the variants of one decision.}
\emph{(a)} Development as a BFS: eight grasps of a mug run at once, and the best
code logic is chosen (decisions 2 and 3 are schematic). \emph{(b)} Of 36
branches, each measured against its parent, few help. \emph{(c)} With more
branches, an accepted change is more likely and larger. \emph{(d)} With more
branches, an accepted change comes sooner in wall-clock time, for more GPU
time.}
\label{fig:branch}
\end{figure}

\finding{4}{Code-Only-as-Policy suits RSI because of its three advantages:
Explicit State, Execution, and Extensibility. Because execution is
controllable, development is also efficient. The variants of a decision can be
developed in parallel, like a Breadth-First Search (BFS).}

\runin{What are important attributes for RSI?} Figure~\ref{fig:rsi} lists four
attributes that we think are important for RSI, and COAP is a better interface
for all four. RSI repeats one loop: find a
failure, change the policy, and keep the change if it helps.

\runin{Efficient: fast evolving through parallel branching.} Development can be
organized as a BFS over decisions. Each node is a decision, such as how to grasp
a mug, and its branches are different ways to make it, such as grasping the mug
by its rim, its handle, or its body. Because every variant is a complete
program that needs no training, the branches are developed in parallel on the
same layouts, and the best code logic is chosen before the search moves on to
the next decision (Figure~\ref{fig:branch}). In comparison, each branch of a
VLA would be a training run, and an Agent Harness calls a large model in every
rollout.

As shown in Figure~\ref{fig:branch}, a single change rarely helps, but with more
branches, a better change is found more often, its gain is larger, and it comes
sooner in wall-clock time, at the cost of more GPU time. Our development
records show the same trend: decisions tried with two or more variants found an
accepted change in four of five cases.

\section{Discussion}
\label{sec:discussion}

This section discusses what the state of Code-Only-as-Policy (COAP) should
contain (\S\ref{sec:tape}), where COAP still fails (\S\ref{sec:errors}), what
sets its upper bound (\S\ref{sec:boundary}), its limitations
(\S\ref{sec:limits}), and how it can serve as a data engine
(\S\ref{sec:dataengine}).

\subsection{What Should the State Contain?}
\label{sec:tape}

The state is the tape of the Turing machine (\S\ref{sec:method}), and its five
categories are shown in Figure~\ref{fig:A} (left). The \textbf{robot state}
describes the robot itself: its joint angles, hand poses, gripper openings, and
what each hand holds. The \textbf{environment state} describes the position,
size, and class of each object, the table, and the objects' spatial relations,
such as which object stands on which. The \textbf{task state} records the task
progress, such as the objects picked so far and the current subgoal. All values
are in physical units and are stored in the program.

\subsection{Error Analysis of COAP}
\label{sec:errors}

\finding{5}{COAP still fails in some episodes. The current coding ability has
limits, and the offline RSI loop needs further optimization so that coding
agents write better code. The failures fall into two groups: limits of state
measurement and limits of the code logic.}

\begin{figure}[!hb]
\centering
\includegraphics[width=\linewidth]{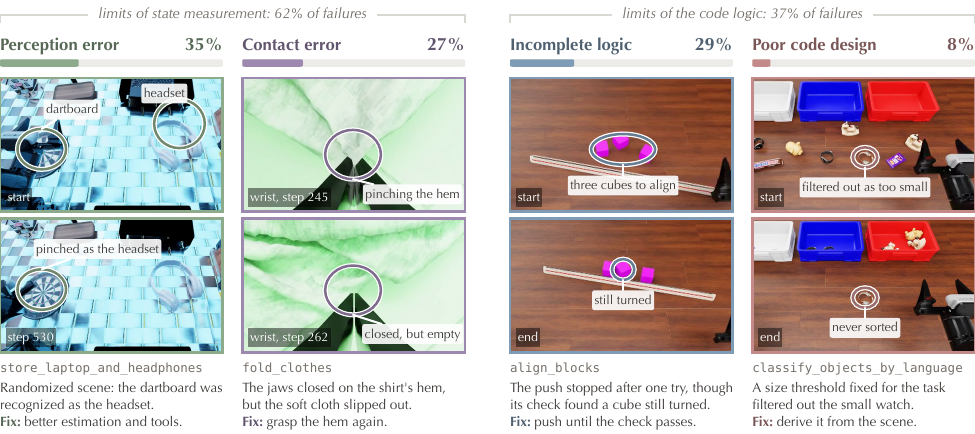}
\caption{\textbf{Four classes of error.} Each column gives the share of
failures in one class, two frames of one failed episode from the official
video, and the fix.}
\label{fig:errors}
\end{figure}

We assigned failed episodes of the runs in Table~\ref{tab:main} to the class of
the smallest change that would have prevented them. Figure~\ref{fig:errors}
shows the share of each class and one example of each.

\runin{Perception error.} The input is RGB only, and recognition is not always
accurate, so a wrong value can enter the state. In Figure~\ref{fig:errors}a, a
dartboard in a randomized scene is recognized as the headset. Such errors may be
more severe in real scenes, but better estimation and tools can help.

\runin{Contact error.} In principle, the more clearly the state is
represented, the higher the upper bound of the code. For contact, especially
with deformable objects, the current representation of the state is coarse and
unclear, so the program cannot always tell how an object is held. Such contact
is inherently hard to represent, but it is still a limit of the state
representation. In Figure~\ref{fig:errors}b, the jaws close on a shirt's hem,
and the soft cloth slips out.

\runin{Incomplete logic.} The code logic is not robust enough, for example
because a retry is missing. In Figure~\ref{fig:errors}c, the push function
stops after one push, although its own check finds a cube still turned. Within
the limited development time, the code was not made fully robust.

\runin{Poor code design.} A function works in the cases it was written for but
does not generalize, for example because of a hard-coded value, such as moving
down by one centimetre. In Figure~\ref{fig:errors}d, a fixed size threshold
filters out a small watch. The fix is to derive the value from the state.

\subsection{The Upper Bound of COAP}
\label{sec:boundary}

\finding{6}{The upper bound of COAP is set by two factors: how accurately and
comprehensively the program represents the state, and how robust
the code logic is, for example whether it is aware of failures and retries.}

\runin{State representation.} Perception and contact errors, 62\% of failures,
are limits of state measurement (Figure~\ref{fig:errors}, left). Perception errors
arise because the input is RGB only, so some estimates are inaccurate. Contact
errors arise because the physical state is hard to represent: for deformable or
otherwise complex objects, part of the contact is not represented well. The
more accurately the state is represented, the higher the upper bound of COAP.

\runin{Robust code logic.} Incomplete logic and poor code design, 37\% of
failures, are limits of the logic (Figure~\ref{fig:errors}, right). More robust
code removes them: functions that notice their own failures, retry with another
method, and derive values from the state.

\subsection{Limitations}
\label{sec:limits}

\runin{Simulation only.} All results are currently in simulation. COAP's perception
relies on the public configuration and assets of the simulated benchmark, such
as the camera parameters, the robot meshes, and the object sizes, which is a
major limitation. On a real robot, these would have to be measured, and
contact-related parameters, such as grip widths and carry speeds, tuned again.

\runin{Offline development.} Online execution is fast, but offline development
takes time: each iteration of the code needs evaluation and revision. Unlike a
model such as GPT, which can be used directly as a policy, COAP has a
development cost for every new task.

\runin{Generalization.} Code may have limited generalization capabilities. In principle, the upper
bound of code is high, and it rises as the state is represented more clearly
(\S\ref{sec:boundary}), but COAP currently succeeds only in 70.24\% of episodes.
Part of the gap comes from incomplete logic and poor code design
(\S\ref{sec:errors}), such as a hard-coded value that works only in the cases
it was written for. In addition, the programs were developed on validation
layouts and never on the test layouts, so lower success on layouts the code
has not seen is to be expected, especially on the randomized layouts of the
Generalization tasks, with new objects, rooms, and distractors (54.1\%, against
77.3\% on their standard layouts).
Still, the code does not depend on the scene: each program is task-level code,
the library is shared across tasks, and on standard layouts, success on test
layouts (74.9\%) stays close to that on validation layouts (76.7\%). Although
code has such limits, in scenes where the state is clear, it runs stably and
can produce data reliably (\S\ref{sec:dataengine}).

\subsection{COAP as an efficient Data Engine for VLAs, Agent-as-Policy and Harness VLA}
\label{sec:dataengine}

\finding{8}{COAP produces successful trajectories quickly and reproducibly, so
it can serve as a data engine for VLAs, Agent-as-Policy, and Harness VLA.}

Once a task program works, it runs on a CPU without a model and can be executed
on many layouts in parallel, so successful trajectories can be obtained very
quickly. Each trajectory comes with the program's explicit state and the
decision behind every action, because the policy is a white box. These
trajectories can train a VLA, and the recorded states and decisions can
supervise the model inside an Agent Harness. COAP can therefore benefit the
development of VLAs and Agent Harnesses.

\runin{A new direction.} Code-Only-as-Policy differs in many ways from VLAs,
World Action Models, and Agent Harnesses (Agent-as-Policy and Harness VLA,
which keep a model in the control loop): its state is explicit, no model runs
during execution, and the policy is improved by coding agents rather than by
training on data. Since its approach differs in so many ways, we view it as a
new direction. If the embodied world is a Turing machine, code can be a
universal interface to the physical world (\S\ref{sec:introduction}). Our
analysis suggests that this direction has advantages
(\S\ref{sec:compare}--\S\ref{sec:dev}), and we hope that our exploration can
benefit VLAs, Agent-as-Policy, Harness VLA, and RSI.

\section{Conclusion}
\label{sec:conclusion}

We propose that the embodied world, a robot together with its environment, is
in essence an Embodied Turing Machine: the world state is its
tape, and the robot policy is its rules. On this basis, we introduce
Code-Only-as-Policy (COAP), whose approach differs from those of existing
directions, such as VLAs, World Action Models, and Agent-as-Policy: the state
is explicit, the rules are code, and coding agents improve the code. On
RoboDojo, COAP succeeds in 70.24\% of episodes, 38.9\% above
the state-of-the-art method, and leads all five dimensions. Compared
with a VLA and an Agent Harness, COAP has three advantages
(\S\ref{sec:compare}): \textbf{Explicit State}, \textbf{Execution}, and
\textbf{Extensibility}. Because the policy is a white box, it suits RSI, and
its development is efficient (\S\ref{sec:dev}). Code still has an upper bound,
set by the state representation and the code logic
(\S\ref{sec:errors}--\S\ref{sec:boundary}), but as coding agents become
stronger, this upper bound will rise. We also believe that COAP can serve as an efficient data engine for VLAs, Agent-as-Policy, and Harness VLA (\S\ref{sec:dataengine}).

\newpage
\appendix
  \startcontents[appendix]
  \printcontents[appendix]{l}{1}{%
    \section*{Appendix Contents}%
    \vspace{-0.5em}%
  }
  \newpage

\definecolor{logicbg}{HTML}{E3ECF3}
\renewcommand{\arraystretch}{1.12}
\section{Advantages of COAP: Details}
\label{app:adv}

\subsection{Execution}
\label{app:exec}

\paragraph{Failure recovery.} Table~\ref{tab:recovery} breaks down the
validation episodes in which something went wrong. \emph{Reattempted}: a failed
step was followed by a new method, such as a re-grasp or a second press.
\emph{Noticed and reported}: the program ended the episode itself and reported
why, such as an object not found or no feasible path.

\begin{table}[h]
\centering
\caption{\textbf{What COAP did when something went wrong.} Left: outcomes of
these episodes. Right: failed steps of the reattempted episodes.}
\label{tab:recovery}
\small
\begin{tabular}{lr@{\hspace{18pt}}lrr}
\toprule
\rowcolor{coapbg}\textbf{Outcome} & \textbf{Share} & \textbf{Failed step (reattempted)} & \textbf{Share} & \textbf{Saved} \\
\midrule
\cellcolor{coapbg!50}Reattempted, succeeded & \cellcolor{coapbg!50}12\% & grasp missed or slipped & 49\% & 17\% \\
Reattempted, failed & 22\% & dropped while carrying & 32\% & 48\% \\
Noticed and reported & 42\% & stuck on release & 18\% & 23\% \\
\quad no further method & 37\% & press did not latch & 8\% & 100\% \\
\quad step budget & 4\% & pour incomplete & 1\% & 100\% \\
Hit the step limit & 1\% & & & \\
Not reported & 22\% & & & \\
\bottomrule
\end{tabular}
\end{table}

\begin{figure}[h]
\centering
\includegraphics[width=\linewidth]{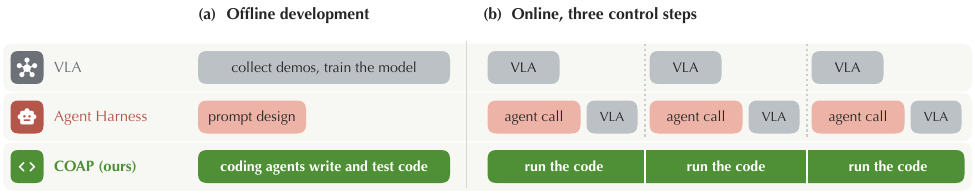}
\caption{\textbf{Compute moves offline.} COAP uses models only offline; a VLA
and an Agent Harness run them online.}
\label{fig:cost}
\end{figure}

\paragraph{Decision time and compute.} The policy makes a median of 352
decisions per episode. The median call returns in 0.3\,ms, because most calls
hand over an action that is already planned; about 25 calls per episode,
perception and planning at the start of a segment, hold 99.4\% of the decision
time. Figure~\ref{fig:B} (right) converts COAP's measured decision time into
compute at $10^{11}$\,FLOP/s, the peak of one CPU core, an upper bound. The
model lines are our estimates from reported model sizes, not measurements: a 3B
VLA that predicts 50-step action chunks ($\pi_0$~\cite{black2025pi0}; we count
one run of about 5\,TFLOP per chunk, which underestimates its compute, since
$\pi_0$ re-plans after 16 or 25 actions), a 7B VLA run every step
(OpenVLA~\cite{kim2024openvla}, about 4\,TFLOP per step), and, for the Agent Harness, one call to a model
with $2\times10^{11}$ active parameters every 100 steps over a context that
starts at 5k tokens and grows by 2k per call. Figure~\ref{fig:cost} shows where
the model compute goes instead.

\paragraph{Success by task length.} Table~\ref{tab:horizon} groups the tasks by
the benchmark's step limit. COAP leads in every group; its success falls on
the longest tasks, as every policy's does, and the strongest Agent Harness
loses most there.

\begin{table}[h]
\centering
\caption{\textbf{Success (\%) by the step limit of a task.} GPT-as-policy: an
Agent Harness without a VLA.}
\label{tab:horizon}
\small
\begin{tabular}{lrrrrr}
\toprule
\rowcolor{coapbg}\textbf{Step limit} & \textbf{Tasks} & \textbf{COAP} & \textbf{Agent Harness} & \textbf{VLA} & \textbf{GPT-as-policy} \\
\midrule
200--500 & 12 & \cellcolor{coapbg!50}\textbf{78.1} & 47.3 & 21.9 & 26.5 \\
550--800 & 15 & \cellcolor{coapbg!50}\textbf{78.4} & 30.6 & 38.0 & 26.9 \\
900--1900 & 15 & \cellcolor{coapbg!50}\textbf{52.0} & 13.2 & 22.1 & 14.8 \\
\bottomrule
\end{tabular}
\end{table}

\subsection{Extensibility}
\label{app:ext}

\begin{figure}[h]
\centering
\includegraphics[width=\linewidth]{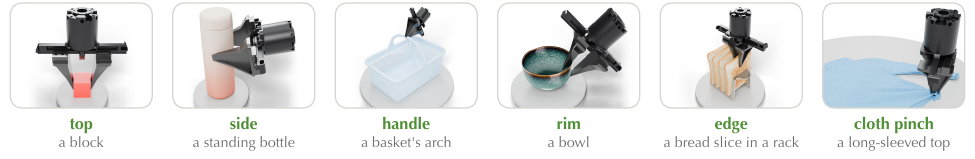}
\caption{\textbf{One example of each grasp method.}}
\label{fig:grasptypes}
\end{figure}

\paragraph{Grasp methods.} Figure~\ref{fig:grasptypes} shows one example of
each grasp method of Figure~\ref{fig:code}b, rendered from the benchmark's
meshes at poses from our grasp code, and Table~\ref{tab:grasptypes} defines
them. A method is decided by the approach and the contact: a grasp that
descends from above is a top grasp even when it leans.

\begin{table}[h]
\centering
\caption{\textbf{Grasp methods.} Tasks: number of task programs using it.}
\label{tab:grasptypes}
\small
\begin{tabular}{llll r}
\toprule
\rowcolor{coapbg}\textbf{Method} & \textbf{Approach} & \textbf{Contact} & \textbf{Examples} & \textbf{Tasks} \\
\midrule
Top & from above & across the body & blocks, digits, a lying pen, eggs & 28 \\
Side & level, toward the axis & around the body or neck & a standing bottle, a cup & 6 \\
Handle & toward the handle & around the bar & a basket's arch, a broom, a mallet & 3 \\
Rim & down onto the wall & one jaw inside, one outside & a bowl, a mug, a plate & 3 \\
Edge & along the thin side & on the two faces & a coin, a bread slice, a key & 4 \\
Cloth pinch & down onto the hem & fabric layers & a long-sleeved top & 1 \\
\bottomrule
\end{tabular}
\end{table}

\paragraph{New tasks.} Figure~\ref{fig:C} (middle) plots the mean best
validation success so far over the four tasks that had no program when the
library was frozen; \texttt{store\_tools\_in\_toolbox}, which never got a
successful program, counts as zero. The two baseline lines are schematic. The
VLA line is a slow ramp over 5--15\,h, the fine-tuning time per task reported
for OpenVLA~\cite{kim2024openvla}, toward the measured average of the strongest
VLA on these tasks (4.7\%). The Agent Harness line
is a straight line with a small slope: its task-specific memory lets it improve,
but slowly, because little carries over from earlier tasks. Table~\ref{tab:newtask}
gives the timeline of each new task.

\begin{table}[h]
\centering
\caption{\textbf{Timeline of the new tasks.} Hours from the start, and
validation success (\%) at each merge.}
\label{tab:newtask}
\small
\setlength{\tabcolsep}{5pt}
\begin{tabular}{lcccc}
\toprule
\rowcolor{coapbg} & \texttt{make\_toast} & \texttt{hang\_mugs} & \texttt{sweep\_blocks} & \texttt{fill\_pen\_holder} \\
\midrule
First evaluation & 3.2\,h & 3.2\,h & 3.9\,h & 0.2\,h \\
First success & 4.7\,h & 5.2\,h & 6.4\,h & 4.0\,h \\
First merge & 5.8\,h & 6.8\,h & 7.9\,h & 2.7\,h \\
\rowcolor{coapbg!50}Success at merges & 50 $\to$ 77 & 7 $\to$ 50 $\to$ 57 & 0 $\to$ 60 $\to$ 50 & 0 $\to$ 20 $\to$ 27 \\
\bottomrule
\end{tabular}
\end{table}

\paragraph{Old tasks.} Between the frozen library and the final code, the old
tasks whose own code did not change kept their success: 78.4\% before and
79.9\% after, on the same layouts (Cochran--Mantel--Haenszel test, $p=0.16$; no
task changed significantly). With the tasks whose code was later cleaned up
included, success went from 80.0\% to 81.4\%. The test detects a pooled shift
of about 4 points, so it rules out a large regression, not every change.

\begin{figure}[h]
\centering
\includegraphics[width=\linewidth]{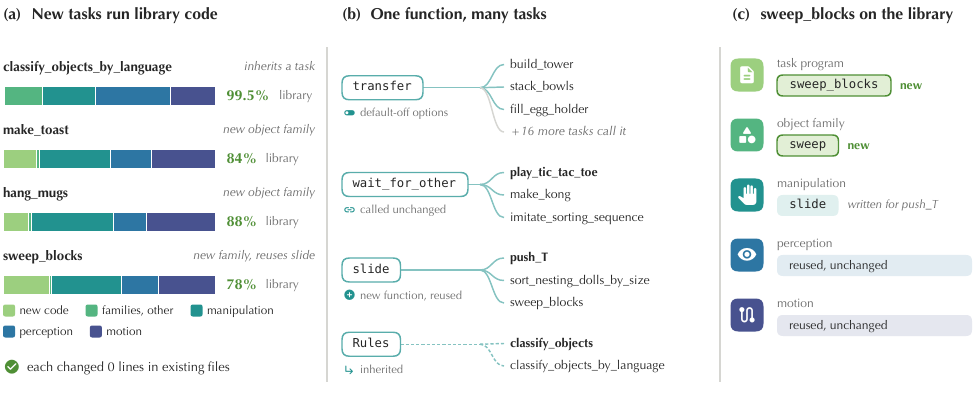}
\caption{\textbf{Extension without breaking.} (a) The code each new task
runs, by layer. (b) One library function reaches many tasks. (c)
\texttt{sweep\_blocks} reuses \texttt{slide}, written for \texttt{push\_T}.}
\label{fig:reuse}
\end{figure}

\paragraph{Extension without breaking.} Figure~\ref{fig:reuse} shows how new
tasks extend the library. A new task adds its own files and changes no line of
an existing one (a). New behaviour reaches other tasks by a call to an existing
library function, by inheritance, or by a parameter that is off by default, so
every other caller runs the same code path as before
(Eq.~\ref{eq:compatibility}) and only the tasks that pass the new parameter
change (b). \texttt{sweep\_blocks}, for example, reuses the sliding function
added to the library for \texttt{push\_T} (c).

\paragraph{Layers.} Table~\ref{tab:layers} lists the layers of the library
(Figure~\ref{fig:code}a). Every task shares manipulation, perception, and
motion, about 14k lines; the families and task programs are larger, but each of
their modules serves few tasks.

\begin{table}[h]
\centering
\caption{\textbf{Layers of the library.} Used by: share of the 40 task
programs that use the layer's most used module.}
\label{tab:layers}
\small
\begin{tabular}{llrr}
\toprule
\rowcolor{coapbg}\textbf{Layer} & \textbf{Holds} & \textbf{Lines} & \textbf{Used by} \\
\midrule
Task programs & order, parameters, special cases & 23.5k & 2.5\% \\
Object families & what an object is, how to hold it & 28.5k & 18\% \\
Manipulation & pick, carry, place, push; grasping & 5.4k & 78\% \\
Perception & look, measure, confirm & 5.3k & 100\% \\
Motion & inverse kinematics, planning, collision & 3.3k & 100\% \\
\bottomrule
\end{tabular}
\end{table}

\section{Parallel Branching: Details}
\label{app:branching}

\paragraph{Branches and the estimate.} A branch is a variant of a program that
changes one decision and is evaluated on the same seeds and layouts as its
parent. Figure~\ref{fig:branch}b uses every branch whose validation success was
measured against its immediate parent. Repeated runs of the same program differ
by one to three episodes per evaluation, so the selection margin of four
episodes is one to two noise widths. The estimate of Figure~\ref{fig:branch}c
draws $k$ effects with replacement from the measured ones and keeps the best if
it reaches the margin, otherwise the parent; the bands are the 5th and 95th
percentiles over bootstrap resamples. The estimate treats branches of one
decision as independent and pools effects across tasks, and each gain is
measured on the layouts that selected it, so the true gain is somewhat lower.

\paragraph{All tasks at once.} In the busiest window of development,
evaluations of 12 tasks ran in parallel and finished in 17\% of the time they
would have taken one after another.

\paragraph{Merges and test.} After the library was frozen, task updates were
merged one at a time, each changing shared code only through default-off
parameters, and overall success rose at every merge, from 66.5\% to 70.24\%.
Changes are selected on validation, and the gains carry over to test: on the 40
tasks with a program, standard success is 76.7\% on validation and 74.9\% on
test, and randomized success 55.2\% and 52.8\%.

\section{Error Analysis: Details}
\label{app:errors}

\paragraph{How a failure is classified.} We read each failed episode from the
start and took the first error that was never recovered. Its class is the
smallest change that would have prevented it, so each failure has exactly one
class (\S\ref{sec:errors}); Table~\ref{tab:errleaves} lists typical cases. The
labels are judgement: a trace shows what the program measured, not what
happened, so we checked the examples against the official videos. Incomplete
logic is judged relative to the robot we have; no fix assumes a new sensor or
actuator.

\begin{table}[H]
\centering
\caption{\textbf{Typical cases of each class of error.}}
\label{tab:errleaves}
\small
\begin{tabular}{ll}
\toprule
\rowcolor{coapbg}\textbf{Class} & \textbf{Typical cases} \\
\midrule
\cellcolor{coapbg!50}Perception error & unrecognized randomized look; wrong final check; split or merged object \\
\cellcolor{coapbg!50}Contact error & slip or drop in the grip; push stops short; tip at release \\
\cellcolor{logicbg}Incomplete logic & no grasp that also reaches the goal; untried order; unchecked collision \\
\cellcolor{logicbg}Poor code design & fixed size filter; fixed grip-width band; fixed match threshold \\
Unknown & no lead in the trace \\
\bottomrule
\end{tabular}
\end{table}

\section{Per-Task Results}
\label{app:pertask}

Table~\ref{tab:pertask} lists the success rate (SR) and the progress score of
COAP on each task.

\begin{table}[h]
\centering
\caption{\textbf{Per-task results of COAP}, by dimension. $^\dagger$Task
without a program.}
\label{tab:pertask}
\small
\setlength{\tabcolsep}{4pt}
\begin{tabular}{lrr@{\hspace{14pt}}lrr}
\toprule
\rowcolor{coapbg}\textbf{Task} & \textbf{SR} & \textbf{Score} & \textbf{Task} & \textbf{SR} & \textbf{Score} \\
\midrule
\multicolumn{3}{l}{\cellcolor{coapbg}\textbf{Memory}} & \multicolumn{3}{l}{\cellcolor{coapbg}\textbf{Generalization}} \\
\texttt{cover\_blocks} & \cellcolor{etmgreen!32}100.0 & 100.0 & \texttt{push\_T} & \cellcolor{etmgreen!31}97.3 & 97.3 \\
\texttt{press\_by\_number} & \cellcolor{etmgreen!32}100.0 & 100.0 & \texttt{stack\_blocks} & \cellcolor{etmgreen!30}94.7 & 95.2 \\
\texttt{swap\_T} & \cellcolor{etmgreen!32}100.0 & 100.0 & \texttt{stack\_bowls} & \cellcolor{etmgreen!30}93.3 & 94.0 \\
\texttt{swap\_blocks} & \cellcolor{etmgreen!32}100.0 & 100.0 & \texttt{fold\_clothes} & \cellcolor{etmgreen!29}90.0 & 91.6 \\
\texttt{imitate\_sorting\_sequence} & \cellcolor{etmgreen!24}74.7 & 78.5 & \texttt{pour\_liquid\_into\_cup} & \cellcolor{etmgreen!29}89.3 & 89.3 \\
\texttt{match\_and\_pick\_from\_conveyor} & \cellcolor{etmgreen!21}64.7 & 64.7 & \texttt{arrange\_largest\_number} & \cellcolor{etmgreen!28}86.7 & 90.0 \\
\multicolumn{3}{l}{\cellcolor{coapbg}\textbf{Precision}} & \texttt{sort\_nesting\_dolls\_by\_size} & \cellcolor{etmgreen!21}64.7 & 64.7 \\
\texttt{play\_Xylophone} & \cellcolor{etmgreen!32}99.3 & 99.3 & \texttt{make\_toast} & \cellcolor{etmgreen!18}56.7 & 60.3 \\
\texttt{build\_tower} & \cellcolor{etmgreen!31}96.7 & 97.3 & \texttt{store\_laptop\_and\_headphones} & \cellcolor{etmgreen!18}55.3 & 77.3 \\
\texttt{insert\_tubes} & \cellcolor{etmgreen!30}94.7 & 96.8 & \texttt{hang\_mugs} & \cellcolor{etmgreen!10}32.0 & 41.9 \\
\texttt{pour\_balls\_into\_vase} & \cellcolor{etmgreen!30}94.0 & 94.0 & \texttt{sweep\_blocks} & \cellcolor{etmgreen!9}28.0 & 28.0 \\
\texttt{fasten\_screws} & \cellcolor{etmgreen!29}91.3 & 95.7 & \texttt{pack\_objects\_into\_box} & \cellcolor{etmgreen!0}0.7 & 6.5 \\
\texttt{deposit\_coin} & \cellcolor{etmgreen!27}85.3 & 86.0 & \multicolumn{3}{l}{\cellcolor{coapbg}\textbf{Long-Horizon}} \\
\texttt{insert\_key} & \cellcolor{etmgreen!13}39.3 & 48.4 & \texttt{play\_tic\_tac\_toe} & \cellcolor{etmgreen!32}99.3 & 99.4 \\
\texttt{plug\_in\_charger}$^\dagger$ & \cellcolor{etmgreen!0}0.0 & -- & \texttt{make\_kong} & \cellcolor{etmgreen!31}98.0 & 98.0 \\
\multicolumn{3}{l}{\cellcolor{coapbg}\textbf{Open}} & \texttt{play\_stacking\_toy} & \cellcolor{etmgreen!26}82.0 & 92.6 \\
\texttt{pour\_by\_language} & \cellcolor{etmgreen!32}100.0 & 100.0 & \texttt{put\_bottles\_into\_dustbin} & \cellcolor{etmgreen!23}73.3 & 82.8 \\
\texttt{align\_blocks} & \cellcolor{etmgreen!30}93.3 & 93.3 & \texttt{classify\_objects} & \cellcolor{etmgreen!16}50.0 & 63.9 \\
\texttt{solve\_equation} & \cellcolor{etmgreen!30}92.7 & 92.7 & \texttt{fill\_egg\_holder} & \cellcolor{etmgreen!8}24.7 & 50.8 \\
\texttt{stack\_blocks\_by\_language} & \cellcolor{etmgreen!29}91.3 & 91.6 & \texttt{fill\_pen\_holder} & \cellcolor{etmgreen!8}24.0 & 49.0 \\
\texttt{general\_pickup} & \cellcolor{etmgreen!20}61.3 & 61.3 & \texttt{organize\_table} & \cellcolor{etmgreen!0}0.0 & 51.5 \\
\texttt{pick\_from\_conveyor\_by\_image} & \cellcolor{etmgreen!15}48.0 & 48.0 &  &  &  \\
\texttt{classify\_objects\_by\_language} & \cellcolor{etmgreen!8}26.0 & 43.9 &  &  &  \\
\texttt{store\_tools\_in\_toolbox}$^\dagger$ & \cellcolor{etmgreen!0}0.0 & -- &  &  &  \\
\bottomrule
\end{tabular}
\end{table}

\section{Case Studies}
\label{app:cases}

\begin{figure}[h]
\centering
\includegraphics[width=\linewidth]{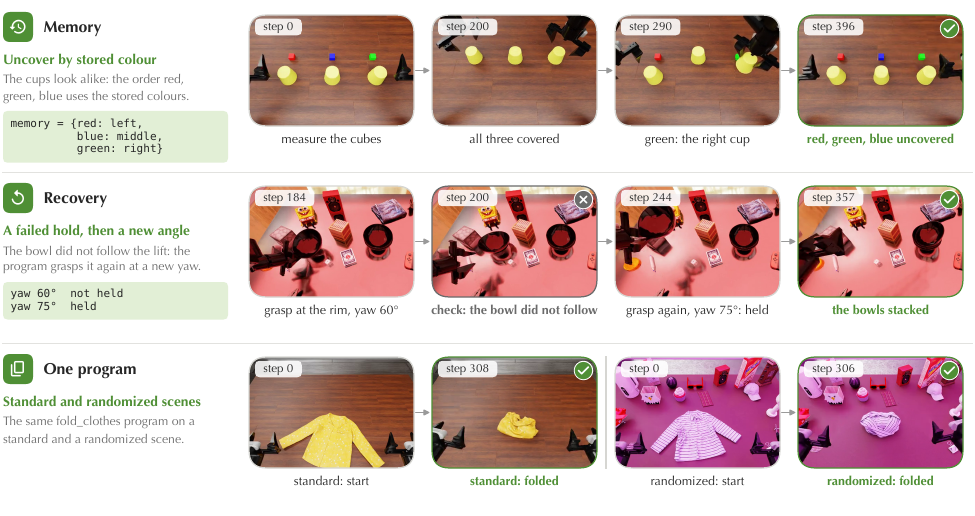}
\caption{\textbf{Case studies:} memory, recovery, and one program in two
scenes.}
\label{fig:cases}
\end{figure}

Figure~\ref{fig:cases} follows three official episodes on development layouts.
In \texttt{cover\_blocks}, the program measures the three cubes once, stores
their colours by position, and uncovers them in the requested order after
identical cups have hidden them (\S\ref{sec:cmp-state}, A2). In a randomized
\texttt{stack\_bowls} episode, the hold check after a rim grasp finds that the
bowl did not follow the lift; the program grasps it again at a new wrist angle
and completes the stack (\S\ref{sec:cmp-exec}, B2). The same
\texttt{fold\_clothes} program folds the shirt in a standard and in a
randomized scene.

\clearpage
\section{Execution Trajectories}
\label{app:trajectories}

\begin{figure}[!htbp]
\centering
\includegraphics[width=0.9\linewidth]{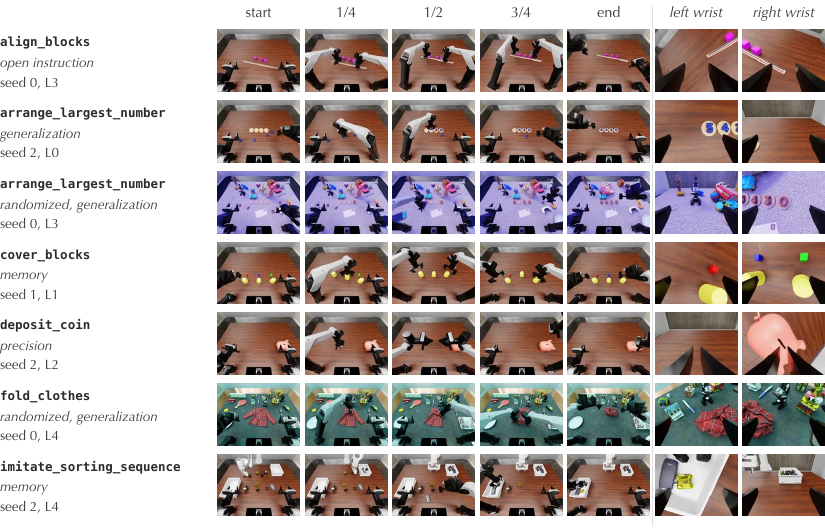}
\caption{\textbf{Execution trajectories (1/4).} Successful official episodes.}
\label{fig:atlas0}
\end{figure}

\begin{figure}[!htbp]
\centering
\includegraphics[width=0.9\linewidth]{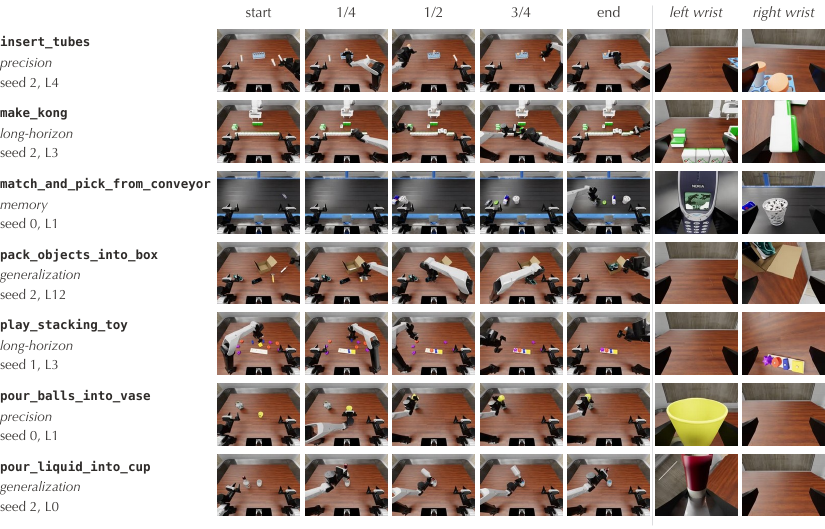}
\caption{\textbf{Execution trajectories (2/4).} Successful official episodes.}
\label{fig:atlas1}
\end{figure}

\begin{figure}[!htbp]
\centering
\includegraphics[width=0.9\linewidth]{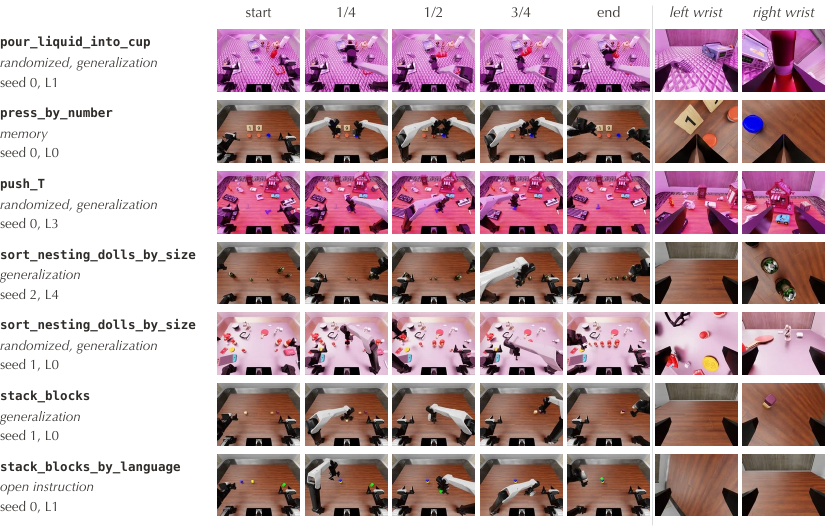}
\caption{\textbf{Execution trajectories (3/4).} Successful official episodes.}
\label{fig:atlas2}
\end{figure}

\begin{figure}[!htbp]
\centering
\includegraphics[width=0.9\linewidth]{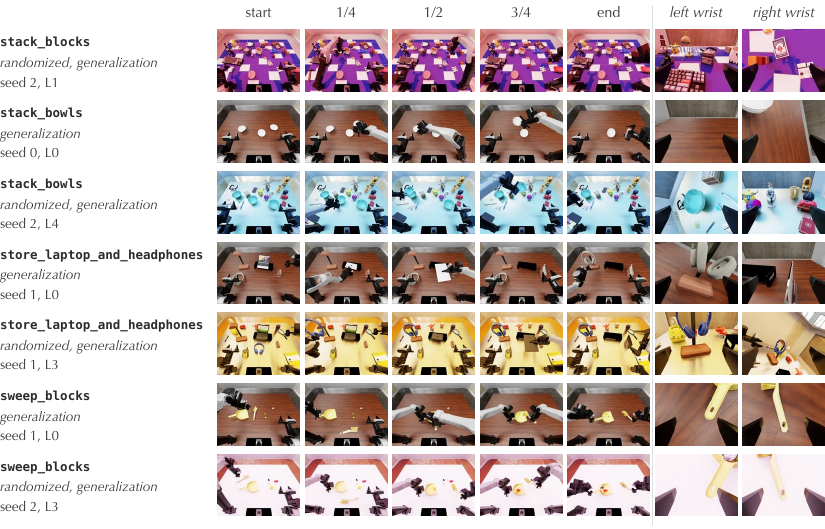}
\caption{\textbf{Execution trajectories (4/4).} Successful official episodes.}
\label{fig:atlas3}
\end{figure}

Figures~\ref{fig:atlas0}--\ref{fig:atlas3} show one successful official
episode per task and variant: the head camera at the start, at one quarter,
half, and three quarters of the episode, and at the end, then the final frames
of the two wrist cameras. Across tasks the same library code recurs: objects
are grasped from above or at a rim, carried along planned paths, and placed
after a close look, while the task program decides their order and targets.

{
\small
\begin{thebibliography}{62}
\providecommand{\natexlab}[1]{#1}
\providecommand{\url}[1]{\texttt{#1}}
\expandafter\ifx\csname urlstyle\endcsname\relax
  \providecommand{\doi}[1]{doi: #1}\else
  \providecommand{\doi}{doi: \begingroup \urlstyle{rm}\Url}\fi

\bibitem[Ahn et~al.(2022)Ahn, Brohan, Brown, Chebotar, Cortes, David, Finn, Fu, et~al.]{ahn2022saycan}
Michael Ahn, Anthony Brohan, Noah Brown, Yevgen Chebotar, Omar Cortes, Byron David, Chelsea Finn, Chuyuan Fu, et~al.
\newblock Do as {I} can, not as {I} say: Grounding language in robotic affordances.
\newblock In \emph{Conference on Robot Learning (CoRL)}, 2022.

\bibitem[{Anthropic}(2025)]{anthropic2025mcp}
{Anthropic}.
\newblock Code execution with {MCP}: Building more efficient agents.
\newblock Anthropic Engineering Blog, \url{https://www.anthropic.com/engineering/code-execution-with-mcp}, 2025.

\bibitem[{Awomo-WAM Team}(2026)]{awomo2026}
{Awomo-WAM Team}.
\newblock {Awomo-0.5}: World action model pre-trained at scale.
\newblock Code repository, \url{https://github.com/Awomo-WestlakeDI/Awomo-0.5}, 2026.

\bibitem[Black et~al.(2025)Black, Brown, Driess, Esmail, Equi, Finn, Fusai, Groom, et~al.]{black2025pi0}
Kevin Black, Noah Brown, Danny Driess, Adnan Esmail, Michael Equi, Chelsea Finn, Niccolo Fusai, Lachy Groom, et~al.
\newblock $\pi_0$: A vision-language-action flow model for general robot control.
\newblock In \emph{Robotics: Science and Systems (RSS)}, 2025.

\bibitem[Brohan et~al.(2023)Brohan, Brown, Carbajal, Chebotar, Chen, Choromanski, Ding, Driess, et~al.]{brohan2023rt2}
Anthony Brohan, Noah Brown, Justice Carbajal, Yevgen Chebotar, Xi~Chen, Krzysztof Choromanski, Tianli Ding, Danny Driess, et~al.
\newblock {RT-2}: Vision-language-action models transfer web knowledge to robotic control.
\newblock In \emph{Conference on Robot Learning (CoRL)}, 2023.

\bibitem[Carion et~al.(2026)Carion, Gustafson, Hu, Debnath, Hu, Suris, Ryali, Alwala, et~al.]{carion2025sam3}
Nicolas Carion, Laura Gustafson, Yuan-Ting Hu, Shoubhik Debnath, Ronghang Hu, Didac Suris, Chaitanya Ryali, Kalyan~Vasudev Alwala, et~al.
\newblock {SAM} 3: Segment anything with concepts.
\newblock In \emph{International Conference on Learning Representations (ICLR)}, 2026.

\bibitem[Chen et~al.(2026{\natexlab{a}})Chen, Chen, Li, Tang, Su, Lu, Wan, Chen, Liu, Yan, et~al.]{chen2026robodojo}
Tianxing Chen, Yue Chen, Zixuan Li, Junyuan Tang, Kailun Su, Haoran Lu, Weijie Wan, Baijun Chen, Songling Liu, Haowen Yan, et~al.
\newblock {RoboDojo}: A unified sim-and-real benchmark for comprehensive evaluation of generalist robot manipulation policies.
\newblock \emph{arXiv preprint arXiv:2607.04434}, 2026{\natexlab{a}}.

\bibitem[Chen et~al.(2026{\natexlab{b}})Chen, Chen, Chen, Cai, Liu, Li, Liang, Lin, et~al.]{chen2025robotwin2}
Tianxing Chen, Zanxin Chen, Baijun Chen, Zijian Cai, Yibin Liu, Zixuan Li, Qiwei Liang, Xianliang Lin, et~al.
\newblock {RoboTwin} 2.0: A scalable data generator and benchmark with strong domain randomization for robust bimanual robotic manipulation.
\newblock In \emph{International Conference on Machine Learning (ICML)}, 2026{\natexlab{b}}.

\bibitem[Chen et~al.(2026{\natexlab{c}})Chen, Bai, Cao, Zeng, Lin, Lin, Liang, Ma, et~al.]{chen2026showharness}
Yanzhe Chen, Zechen Bai, Zhijun Cao, Wenzheng Zeng, Kevin~Qinghong Lin, Yiqi Lin, Guoqiang Liang, Kevin~Yuchen Ma, et~al.
\newblock {Show-Harness}: Just a {VLM} agent can play robots.
\newblock \emph{arXiv preprint arXiv:2609.10522}, 2026{\natexlab{c}}.

\bibitem[{Dexmal}(2026)]{dm052026}
{Dexmal}.
\newblock {DM0.5}: Designed for the open world, where generalization emerges.
\newblock Blog post, \url{https://www.dexmal.com/blog/dm0.5/index_en.html}, 2026.

\bibitem[Elmaaroufi et~al.(2026)Elmaaroufi, Svegliato, Kalade, Schelle, Seshia, and Zaharia]{elmaaroufi2026rho}
Karim Elmaaroufi, Justin Svegliato, Sarunas Kalade, Graham Schelle, Sanjit~A. Seshia, and Matei Zaharia.
\newblock {RHO}: Your coding agent is secretly a roboticist.
\newblock \emph{arXiv preprint arXiv:2606.16458}, 2026.

\bibitem[Fu et~al.(2026)Fu, Yu, El-Refai, Kou, Xue, Huang, Xiao, et~al.]{fu2026capx}
Letian Fu, Justin Yu, Karim El-Refai, Ethan Kou, Haoru Xue, Huang Huang, Wenli Xiao, et~al.
\newblock {CaP-X}: A framework for benchmarking and improving coding agents for robot manipulation.
\newblock In \emph{International Conference on Machine Learning (ICML)}, 2026.

\bibitem[{Futian Lab and VisIncept}(2026)]{rexm12026}
{Futian Lab and VisIncept}.
\newblock {Rex-M1 Preview}.
\newblock RoboDojo policy adapter, \url{https://github.com/XPolicyLab/XPolicyLab/pull/131}, 2026.

\bibitem[He et~al.(2026)He, Li, Wang, Yu, Zhu, He, Liu, Zhang, et~al.]{he2026mebrain}
Wei He, Hengtao Li, Chenfeng Wang, Zhongrui Yu, Xuhan Zhu, Maokui He, Zide Liu, Xiyue Zhang, et~al.
\newblock {ME-Brain-1.0:} Memory, cognition and action for evolving embodied intelligence.
\newblock \emph{arXiv preprint arXiv:2609.24271}, 2026.

\bibitem[{HKU MMLab and Kinetix AI}(2026)]{physicalrsi2026}
{HKU MMLab and Kinetix AI}.
\newblock {PhysicalRSI 1.0}: Recursive self-harness for scaling embodied skills.
\newblock Project page, \url{https://mmlab.hk/research/PhysicalRSI}, 2026.

\bibitem[Huang et~al.(2022)Huang, Xia, Xiao, Chan, Liang, Florence, Zeng, Tompson, et~al.]{huang2022inner}
Wenlong Huang, Fei Xia, Ted Xiao, Harris Chan, Jacky Liang, Pete Florence, Andy Zeng, Jonathan Tompson, et~al.
\newblock Inner monologue: Embodied reasoning through planning with language models.
\newblock In \emph{Conference on Robot Learning (CoRL)}, 2022.

\bibitem[Huang et~al.(2023)Huang, Wang, Zhang, Li, Wu, and Fei-Fei]{huang2023voxposer}
Wenlong Huang, Chen Wang, Ruohan Zhang, Yunzhu Li, Jiajun Wu, and Li~Fei-Fei.
\newblock {VoxPoser}: Composable {3D} value maps for robotic manipulation with language models.
\newblock In \emph{Conference on Robot Learning (CoRL)}, 2023.

\bibitem[Kim et~al.(2026)Kim, Jang, Koo, Jang, Kim, Kim, Yoon, Jang, et~al.]{kim2026rldx}
Dongyoung Kim, Huiwon Jang, Myungkyu Koo, Suhyeok Jang, Taeyoung Kim, Beomjun Kim, Byungjun Yoon, Changsung Jang, et~al.
\newblock {RLDX-1} technical report.
\newblock \emph{arXiv preprint arXiv:2605.03269}, 2026.

\bibitem[Kim et~al.(2024)Kim, Pertsch, Karamcheti, Xiao, Balakrishna, Nair, Rafailov, Foster, et~al.]{kim2024openvla}
Moo~Jin Kim, Karl Pertsch, Siddharth Karamcheti, Ted Xiao, Ashwin Balakrishna, Suraj Nair, Rafael Rafailov, Ethan Foster, et~al.
\newblock {OpenVLA}: An open-source vision-language-action model.
\newblock In \emph{Conference on Robot Learning (CoRL)}, 2024.

\bibitem[Li et~al.(2026)Li, Song, Zhao, Wang, Ding, Wang, Zeng, and Li]{li2025spatialforcing}
Fuhao Li, Wenxuan Song, Han Zhao, Jingbo Wang, Pengxiang Ding, Donglin Wang, Long Zeng, and Haoang Li.
\newblock Spatial {Forcing:} Implicit spatial representation alignment for vision-language-action model.
\newblock In \emph{International Conference on Learning Representations (ICLR)}, 2026.

\bibitem[Liang et~al.(2023)Liang, Huang, Xia, Xu, Hausman, Ichter, Florence, and Zeng]{liang2023code}
Jacky Liang, Wenlong Huang, Fei Xia, Peng Xu, Karol Hausman, Brian Ichter, Pete Florence, and Andy Zeng.
\newblock Code as policies: Language model programs for embodied control.
\newblock In \emph{IEEE International Conference on Robotics and Automation (ICRA)}, 2023.

\bibitem[{LiberAI}(2026)]{liber2026lite}
{LiberAI}.
\newblock {Liber-0 Lite}.
\newblock RoboDojo policy adapter, \url{https://github.com/XPolicyLab/XPolicyLab/tree/main/policy/liber_0_lite}, 2026.

\bibitem[{LiberAI}(2026)]{liber2026preview}
{LiberAI}.
\newblock {Liber-0 Preview}.
\newblock RoboDojo leaderboard entry, \url{https://robodojo-benchmark.com/LeaderBoard}, 2026.

\bibitem[Liu et~al.(2023)Liu, Zhu, Gao, Feng, Liu, Zhu, and Stone]{liu2023libero}
Bo~Liu, Yifeng Zhu, Chongkai Gao, Yihao Feng, Qiang Liu, Yuke Zhu, and Peter Stone.
\newblock {LIBERO}: Benchmarking knowledge transfer for lifelong robot learning.
\newblock In \emph{Advances in Neural Information Processing Systems (NeurIPS), Datasets and Benchmarks Track}, 2023.

\bibitem[Liu et~al.(2026)Liu, Dong, Ye, Yuan, Jiang, Yang, Cao, Liu, et~al.]{galaxea2026g05}
Yicheng Liu, Zibin Dong, Baijun Ye, Tianyuan Yuan, Tao Jiang, Anqi Yang, Shicheng Cao, Haonan Liu, et~al.
\newblock {G0.5:} One autoregressive stream for robot reasoning and action.
\newblock \emph{arXiv preprint arXiv:2608.11739}, 2026.

\bibitem[Lu et~al.(2026)Lu, Wu, Kou, Fu, Xiao, Mandlekar, Xu, Shi, et~al.]{lu2026aspire}
Runyu Lu, Yubo Wu, Ethan Kou, Letian Fu, Wenli Xiao, Ajay Mandlekar, Yinzhen Xu, Guanya Shi, et~al.
\newblock {ASPIRE}: Agentic /skills discovery for robotics.
\newblock \emph{arXiv preprint arXiv:2607.00272}, 2026.

\bibitem[Ma et~al.(2026)Ma, Cai, Xu, Li, Yang, Tian, Cao, Zhu, et~al.]{ma2026internvla}
Haoxiang Ma, Junhao Cai, Xiaoxu Xu, Hao Li, Yuyin Yang, Yang Tian, Jiafei Cao, Hongrui Zhu, et~al.
\newblock {InternVLA-A1.5:} Unifying understanding, latent foresight, and action for compositional generalization.
\newblock \emph{arXiv preprint arXiv:2607.04988}, 2026.

\bibitem[{Meituan Robotics}(2026)]{meituan2026robotics0}
{Meituan Robotics}.
\newblock {Meituan-Robotics-0}.
\newblock RoboDojo policy adapter, \url{https://github.com/XPolicyLab/XPolicyLab/tree/main/policy/Meituan_Robotics_0}, 2026.

\bibitem[Miao et~al.(2026)Miao, Li, Fang, Song, Wang, Zhang, Wang, Li, et~al.]{miao2026internw0}
Xingyu Miao, Zizun Li, Baole Fang, Kaiwen Song, Tenghui Wang, Hanxue Zhang, Yating Wang, Xudong Li, et~al.
\newblock {InternW0-$\Delta$:} A world action model bridging predictive dynamics and actions with {20K+} hours of open data.
\newblock \emph{arXiv preprint arXiv:2609.31394}, 2026.

\bibitem[Naouali et~al.(2026)Naouali, Wu, Wong, Puthran, and Younis]{naouali2026vlcp}
Dhia Naouali, Minghan Wu, Claudia Wong, Abhinav Puthran, and Omar~G. Younis.
\newblock {VLCP}: Vision language control policy closed-loop code replanning for robot manipulation.
\newblock \emph{arXiv preprint arXiv:2608.16978}, 2026.

\bibitem[Nasiriany et~al.(2026)Nasiriany, Nasiriany, Maddukuri, and Zhu]{nasiriany2026robocasa365}
Soroush Nasiriany, Sepehr Nasiriany, Abhiram Maddukuri, and Yuke Zhu.
\newblock {RoboCasa365}: A large-scale simulation framework for training and benchmarking generalist robots.
\newblock In \emph{International Conference on Learning Representations (ICLR)}, 2026.

\bibitem[{Noematrix Team}(2026)]{noematrix2026roborsi}
{Noematrix Team}.
\newblock {RoboRSI}: Stable, efficient, and reusable robot self-evolution in complex real-world environments.
\newblock Blog post, \url{https://lab.noematrix.ai/blog/2-roborsi/}, 2026.

\bibitem[{OpenAI}(2025)]{openai2025codexcli}
{OpenAI}.
\newblock Codex {CLI}.
\newblock \url{https://github.com/openai/codex}, 2025.

\bibitem[{OpenAI}(2026{\natexlab{a}})]{openai2026astra}
{OpenAI}.
\newblock {GPT-6 Astra}: A new generation of intelligence.
\newblock \url{https://openai.com/index/gpt-6-astra/}, 2026{\natexlab{a}}.

\bibitem[{OpenAI}(2026{\natexlab{b}})]{openai2026pricing}
{OpenAI}.
\newblock Pricing.
\newblock OpenAI API documentation, \url{https://developers.openai.com/api/docs/pricing}, 2026{\natexlab{b}}.

\bibitem[{Physical Intelligence} et~al.(2025){Physical Intelligence}, Black, Brown, Darpinian, Dhabalia, Driess, Esmail, Equi, et~al.]{intelligence2025pi05}
{Physical Intelligence}, Kevin Black, Noah Brown, James Darpinian, Karan Dhabalia, Danny Driess, Adnan Esmail, Michael Equi, et~al.
\newblock $\pi_{0.5}$: a vision-language-action model with open-world generalization.
\newblock In \emph{Conference on Robot Learning (CoRL)}, 2025.

\bibitem[{RobotEra}(2026)]{vpp22026}
{RobotEra}.
\newblock {VPP2-Preview}.
\newblock RoboDojo leaderboard entry, \url{https://robodojo-benchmark.com/LeaderBoard}, 2026.

\bibitem[Schick et~al.(2023)Schick, Dwivedi-Yu, Dess{\`\i}, Raileanu, Lomeli, Hambro, Zettlemoyer, Cancedda, and Scialom]{schick2023toolformer}
Timo Schick, Jane Dwivedi-Yu, Roberto Dess{\`\i}, Roberta Raileanu, Maria Lomeli, Eric Hambro, Luke Zettlemoyer, Nicola Cancedda, and Thomas Scialom.
\newblock Toolformer: Language models can teach themselves to use tools.
\newblock In \emph{Advances in Neural Information Processing Systems (NeurIPS)}, 2023.

\bibitem[Shen et~al.(2026)Shen, You, Liu, Zheng, Zha, Yamazaki, Zhang, Huang, et~al.]{shen2026embodiedswe}
Zeyu Shen, Haoxiang You, Yilang Liu, Zhicheng Zheng, Lihan Zha, Kashu Yamazaki, Mingtong Zhang, Suning Huang, et~al.
\newblock {EmbodiedSWE}: Coding agents for long horizon dexterous robotics.
\newblock \emph{arXiv preprint arXiv:2609.27308}, 2026.

\bibitem[{Simate}(2026)]{simate2026}
{Simate}.
\newblock {Simate-beta}.
\newblock RoboDojo policy adapter, \url{https://github.com/XPolicyLab/XPolicyLab/tree/main/policy/Simate_beta}, 2026.

\bibitem[Singh et~al.(2023)Singh, Blukis, Mousavian, Goyal, Xu, Tremblay, Fox, Thomason, et~al.]{singh2023progprompt}
Ishika Singh, Valts Blukis, Arsalan Mousavian, Ankit Goyal, Danfei Xu, Jonathan Tremblay, Dieter Fox, Jesse Thomason, et~al.
\newblock {ProgPrompt}: Generating situated robot task plans using large language models.
\newblock In \emph{IEEE International Conference on Robotics and Automation (ICRA)}, 2023.

\bibitem[{StarVLA Community}(2026)]{starvla2026codebase}
{StarVLA Community}.
\newblock {StarVLA}: A lego-like codebase for vision-language-action model developing.
\newblock \emph{arXiv preprint arXiv:2604.05014}, 2026.

\bibitem[Su et~al.(2026)Su, Zheng, Yan, Yi, Zhang, and Wang]{su2026astra}
Jiayi Su, Yixin Zheng, Mi~Yan, Li~Yi, Zhizheng Zhang, and He~Wang.
\newblock {GPT 6 Astra} as an embodied policy.
\newblock Technical report, \url{https://anonymous-report-421.github.io/public-website/}, 2026.

\bibitem[Varda and Pai(2025)]{varda2025codemode}
Kenton Varda and Sunil Pai.
\newblock {Code Mode}: the better way to use {MCP}.
\newblock Cloudflare Blog, \url{https://blog.cloudflare.com/code-mode/}, 2025.

\bibitem[Wang et~al.(2024{\natexlab{a}})Wang, Xie, Jiang, Mandlekar, Xiao, Zhu, Fan, and Anandkumar]{wang2023voyager}
Guanzhi Wang, Yuqi Xie, Yunfan Jiang, Ajay Mandlekar, Chaowei Xiao, Yuke Zhu, Linxi Fan, and Anima Anandkumar.
\newblock Voyager: An open-ended embodied agent with large language models.
\newblock \emph{Transactions on Machine Learning Research (TMLR)}, 2024{\natexlab{a}}.

\bibitem[Wang et~al.(2024{\natexlab{b}})Wang, Chen, Yuan, Zhang, Li, Peng, and Ji]{wang2024codeact}
Xingyao Wang, Yangyi Chen, Lifan Yuan, Yizhe Zhang, Yunzhu Li, Hao Peng, and Heng Ji.
\newblock Executable code actions elicit better {LLM} agents.
\newblock In \emph{International Conference on Machine Learning (ICML)}, 2024{\natexlab{b}}.

\bibitem[Wang et~al.(2026)Wang, Huang, Li, Zhang, Liang, Jin, Chen, Chi, et~al.]{wang2026openwam}
Yuran Wang, Siqiao Huang, Mingleyang Li, Chenhao Zhang, Jiaqi Liang, Weiyang Jin, Yue Chen, Xuemin Chi, et~al.
\newblock {OpenWAM:} An open, modular exploration towards systematic world-action model pretraining.
\newblock \emph{arXiv preprint arXiv:2609.07398}, 2026.

\bibitem[Wu et~al.(2026)Wu, Lu, Wang, Yang, Liu, Wang, Zhu, Sun, et~al.]{wu2026pragmatic}
Wei Wu, Fan Lu, Yunnan Wang, Shuai Yang, Shi Liu, Fangjing Wang, Qian Zhu, He~Sun, et~al.
\newblock A pragmatic {VLA} foundation model.
\newblock \emph{arXiv preprint arXiv:2601.18692}, 2026.

\bibitem[{Xiaomi Robotics Team} et~al.(2026){Xiaomi Robotics Team}, Guo, Jin, Li, Li, Li, Liu, Peng, et~al.]{xiaomi2026robotics1}
{Xiaomi Robotics Team}, Jun Guo, Piaopiao Jin, Jason Li, Peiyan Li, Yingyan Li, Futeng Liu, Wanli Peng, et~al.
\newblock {Xiaomi-Robotics-1:} Scaling vision-language-action models with over {100K} hours of real-world trajectories.
\newblock \emph{arXiv preprint arXiv:2607.15330}, 2026.

\bibitem[Yang et~al.(2024)Yang, Jimenez, Wettig, Lieret, Yao, Narasimhan, and Press]{yang2024sweagent}
John Yang, Carlos~E. Jimenez, Alexander Wettig, Kilian Lieret, Shunyu Yao, Karthik Narasimhan, and Ofir Press.
\newblock {SWE-agent}: Agent-computer interfaces enable automated software engineering.
\newblock In \emph{Advances in Neural Information Processing Systems (NeurIPS)}, 2024.

\bibitem[Yang et~al.(2026{\natexlab{a}})Yang, Wang, Chen, Wang, Tang, Gui, Ye, Lu, et~al.]{yang2026vlact}
Senqiao Yang, Chengyao Wang, Yuxin Chen, Zixuan Wang, Longxiang Tang, Haokun Gui, Jinhui Ye, Changsheng Lu, et~al.
\newblock Beyond data scaling: Representation-centric continued pre-training for vision-language-action models.
\newblock \emph{arXiv preprint arXiv:2608.27550}, 2026{\natexlab{a}}.

\bibitem[Yang et~al.(2026{\natexlab{b}})Yang, Zheng, Wang, Kou, Li, and Yang]{yang2026kinrt}
Tianhang Yang, Yanze Zheng, Junjie Wang, Wei-Bin Kou, Ruotong Li, and Yujiu Yang.
\newblock Route by kinematics, act by observation: Kinematics-Supervised expert routing in {MoE-Augmented} {VLA}.
\newblock \emph{arXiv preprint arXiv:2607.26807}, 2026{\natexlab{b}}.

\bibitem[Yao et~al.(2023)Yao, Zhao, Yu, Du, Shafran, Narasimhan, and Cao]{yao2023react}
Shunyu Yao, Jeffrey Zhao, Dian Yu, Nan Du, Izhak Shafran, Karthik Narasimhan, and Yuan Cao.
\newblock {ReAct}: Synergizing reasoning and acting in language models.
\newblock In \emph{International Conference on Learning Representations (ICLR)}, 2023.

\bibitem[Yin et~al.(2026)Yin, Xu, Yang, Tam, Liu, Yao, Zeng, Cui, et~al.]{yin2026simplememvla}
Cheng Yin, Wang Xu, Junpeng Yang, Sikyuen Tam, Hanyu Liu, Yuan Yao, Xiangrui Zeng, Junbo Cui, et~al.
\newblock {SimpleMemVLA:} A simple but effective native-video memory for vision-language-action models.
\newblock \emph{arXiv preprint arXiv:2609.05533}, 2026.

\bibitem[Zhang et~al.(2026)Zhang, Zhang, Gao, Li, Liu, Nie, Zhu, Qiu, et~al.]{zhang2026harness}
Yixian Zhang, Huanming Zhang, Feng Gao, Xiao Li, Zhihao Liu, Yi~Nie, Chunyang Zhu, Jiaxing Qiu, et~al.
\newblock Harness {VLA}: Steering frozen {VLAs} into reliable manipulation primitives via memory-guided agents.
\newblock \emph{arXiv preprint arXiv:2607.08448}, 2026.

\bibitem[Zhang et~al.(2026)Zhang, Wang, Ouyang, Huang, Li, Su, Jin, Chai, et~al.]{zhang2026astra}
Wenbo Zhang, Kaixuan Wang, Yutao Ouyang, Xiaoyu Huang, Liyang Li, Kailun Su, Weiyang Jin, Wenhao Chai, et~al.
\newblock An unexpected robot policy: Early evaluations of {GPT-6} {Astra} on {RoboDojo} and beyond.
\newblock \emph{arXiv preprint arXiv:2609.24170}, 2026.

\bibitem[Zhang et~al.(2026)Zhang, Xiang, Lin, Huang, Wang, Zhong, Dong, Wu, et~al.]{zhang2026hyembodied}
He Zhang, Lingzhu Xiang, Haitao Lin, Zeyu Huang, Minghui Wang, Dingyan Zhong, Yubo Dong, Yihao Wu, et~al.
\newblock {Hy-Embodied-0.5-VLA:} From vision-language-action models to a real-world robot learning stack.
\newblock \emph{arXiv preprint arXiv:2606.14409}, 2026.

\bibitem[Zhou et~al.(2025)Zhou, Xu, Tie, Chen, Zhang, Chu, Zhou, and Sun]{zhou2025liberopro}
Xueyang Zhou, Yangming Xu, Guiyao Tie, Yongchao Chen, Guowen Zhang, Duanfeng Chu, Pan Zhou, and Lichao Sun.
\newblock {LIBERO-PRO}: Towards robust and fair evaluation of vision-language-action models beyond memorization.
\newblock \emph{arXiv preprint arXiv:2510.03827}, 2025.

\bibitem[Si et~al.(2026)Si, Bi, Yang, Ni, Huang, Wang, Jiang, Wang, et~al.]{si2026pyrua}
Ruiyang Si, Jianxin Bi, Shunyu Yang, Rui Ni, Wenbo Huang, Qiang Wang,
Shulong Jiang, Duomin Wang, et~al.
\newblock Fewer tokens, better action: {GPT-6 Astra} robot agents with 14\%
higher success rate but 65\% fewer tokens.
\newblock \emph{arXiv preprint arXiv:2610.01939}, 2026.

\bibitem[Turing(1936)]{turing1936computable}
Alan M. Turing.
\newblock On computable numbers, with an application to the
{Entscheidungsproblem}.
\newblock \emph{Proceedings of the London Mathematical Society},
s2-42(1):230--265, 1936.

\bibitem[Zhu et~al.(2026)Zhu, Hong, Sun, Xu, He, Chen, Wang, Yuan, et~al.]{zhu2026vla}
Jiarun Zhu, Yijun Hong, Xiaoquan Sun, Zetian Xu, Qijun He, Haijier Chen, Zhiyong Wang, Mingqi Yuan, et~al.
\newblock Can vision-language-action models learn from real-world data continually without forgetting?
\newblock \emph{arXiv preprint arXiv:2605.26820}, 2026.

\bibitem[He et~al.(2026)He, Ma, Yu, Chen, Huang, Wu, Yang, and Jiang]{he2026skill2real}
Xincheng He, Siyu Ma, Chang Yu, Yunuo Chen, Yanjia Huang, Ying~Nian Wu, Yin Yang, and Chenfanfu Jiang.
\newblock {Skill2Real}: Agentic skill learning for zero-shot sim-to-real robot manipulation.
\newblock \emph{arXiv preprint arXiv:2610.02788}, 2026.

\end{thebibliography}

}
\end{document}